\pdfoutput=1

\documentclass[11pt]{article}

\usepackage[final]{acl}

\usepackage{times}
\usepackage{latexsym}
\usepackage{comment}

\usepackage[T1]{fontenc}

\usepackage[utf8]{inputenc}

\usepackage{microtype}

\usepackage{inconsolata}

\usepackage{graphicx}
\usepackage{amsmath}
\usepackage{amssymb}
\usepackage{mathtools}
\usepackage{subfig}
\usepackage{booktabs} 
\usepackage[table]{xcolor}
\usepackage{xcolor,colortbl}
\usepackage{xurl}

\title{Are you sure? Measuring models bias in content moderation through uncertainty}

\author{
 \textbf{Alessandra Urbinati\textsuperscript{1}},
 \textbf{Mirko Lai\textsuperscript{2,3}},
 \textbf{Simona Frenda\textsuperscript{4,3}},
  \textbf{Marco Antonio Stranisci\textsuperscript{5,3}}
\\
\\
 \textsuperscript{1}Laboratory for the Modeling of Biological and Socio-technical Systems, Northeastern \\University, Boston, MA, USA,
 \textsuperscript{2}Heriot-Watt University, Edinburgh, Scotland,
 \\
  \textsuperscript{3}aequa-tech, Torino, Italy,
  \textsuperscript{4}Università del Piemonte Orientale, Vercelli, Italy,
  \\
  \textsuperscript{5}Università degli Studi di Torino, Torino, Italy
  \\
  \small{
   \textbf{Correspondence:} \href{mailto:marcoantonio.stranisci@unito.it}{marcoantonio.stranisci@unito.it} 
  }
}

\begin{document}
\maketitle
\begin{abstract}
Automatic content moderation is crucial to ensuring safety in social media. Language Model-based classifiers are being increasingly adopted for this task, but it has been shown that they perpetuate racial and social biases. Even if several resources and benchmark corpora have been developed to challenge this issue, measuring the fairness of models in content moderation remains an open issue. In this work, we present an unsupervised approach that benchmarks models on the basis of their uncertainty in classifying messages annotated by people belonging to vulnerable groups.
We use uncertainty, computed by means of the conformal prediction technique, as a proxy to analyze the bias of 11 models against women and non-white annotators and observe to what extent it diverges from metrics based on performance, such as the $F_1$ score. The results show that some pre-trained models predict with high accuracy the labels coming from minority groups, even if the confidence in their prediction is low.
Therefore, by measuring the confidence of models, we are able to see which groups of annotators are better represented in pre-trained models and lead the debiasing process of these models before their effective use.\footnote{Code and data of our experiments are on \url{https://github.com/aequa-tech/conformal-prediction-bias}}
\end{abstract}

\section{Introduction}
\label{intro}

Language models (LMs), including large language models (LLMs), have been widely used in real-world applications since their introduction due to their ability to generate and understand human-like text and have been adopted as street-level algorithms \cite{alkhatib2019street}: technologies that are implemented to enforce the rules of platforms based on user-generated content. In such a sense, these technologies play the role of bureaucrats, as they provide an interpretation of community guidelines and decide accordingly which users' behaviors must be removed and which not. Street-level algorithms are widely present in social media \cite{jiang2020reasoning}, but their adoption is not limited to this context. Toxicity classifiers such as the Perspective API \cite{hosseini2017deceiving} are used to filter out unwanted texts from pretraining data, playing a crucial role in the development of fair 
models.
\citet{dignum2023responsible} highlights 
the over-reliance on data of these technologies,
leaving the power of what the models know and 
how to address the problems
to those who produce and maintain the data. 
The societal cost of doing that is high, and it has been demonstrated
that such technologies tend to perpetuate social biases against vulnerable minorities, 
whose opinion is less represented or excluded in data \cite{kalluri2020don}.


In this paper, we investigate the presence of biases towards underrepresented groups in hate speech detection, 
using two datasets, Social Bias Inference Corpus \cite{sap2020social} and CREHate \cite{lee-etal-2024-exploring-cross}, and across two dimensions: gender and ethnicity. As \textit{bias}, we refer to systematic discrimination against a disadvantaged group of people \cite{friedman1996bias}, who will suffer from \textit{representational harms} because systems tend to fail to recognize their existence \cite{wang2022measuring}. 
We think that measuring the uncertainty of models \textit{through the lens} of annotators with different sociodemographics 
could help to identify social biases against vulnerable groups.

To this end, we exploit the conformal prediction framework to assess the uncertainty and reliability of model predictions. Unlike conventional approaches that prioritize accuracy, conformal prediction provides a metric for evaluating the alignment between model outputs and the confidence required for fair decision-making. By identifying disparities, this framework offers a structured approach to understanding 
biases, ultimately fostering fairness and inclusivity in model design and evaluation.

Specifically, we formulate two research questions.

\textbf{RQ 1. 
Is a models' uncertainty in automatic content moderation a predictor of biases against vulnerable groups to discrimination?} 

\textbf{RQ 2. Can the fairness of models be assessed using user representations based on uncertainty?} 

Addressing these questions, we show that our approach brings out general patterns of hidden discrimination against non-white people but also shows that some street-level algorithms are expected to be fairer than others.

The main contributions of this research are the following: \textit{i.} We introduce an unsupervised approach that leverages uncertainty to assess the fairness of 
models' predictions; \textit{ii.} We provide a benchmarking analysis of $11$ NLP systems that exhibit different levels of alignment with annotations provided by annotators belonging to vulnerable groups to discrimination; \textit{iii.} We demonstrate that representing users through the uncertainty of model predictions is effective to observe the tendency of models to align with specific socio-demographic groups.


\section{Related Work}
\label{related_work}

Unlike structured domains where uncertainty estimation is often a standard consideration, NLP research has traditionally focused on maximizing accuracy-based evaluation metrics, such as $F_1$ score and log-likelihood,
underestimating the computation of model uncertainty. Uncertainty measures reached, only recently, visibility in the NLP community \cite{Xiao_Wang_2019,uncertainlp-2024-uncertainty}. 
NLP tasks, indeed, 
involve inherently ambiguous data, where 
human label variation
introduces significant variability into annotated datasets. 
And, in such cases, conventional evaluation metrics fail to capture the full extent of model uncertainty, potentially leading to overconfident and unreliable predictions.

\paragraph{Bias detection.}
Scientific literature has revealed important biases coming from data created 
and annotated by specific segments of the population, leading to the creation of non-neutral models \cite{santurkar2023whose} and to the reinforcement of social stereotypes \cite{caliskan2017semantics}.
Various techniques have been proposed to reveal biases and models' viewpoints:
evaluation of contextualized word embeddings \cite{basta-etal-2019-evaluating,ethayarajh-etal-2019-understanding}, 
specific evaluation frameworks \cite{barikeri-etal-2021-redditbias,felkner-etal-2023-winoqueer}, 
questionnaires \cite{scherrer2024evaluating,wright2024llm}, 
transformer-based recognizers \cite{DBLP:journals/corr/abs-2312-10075}, 
special prompts \cite{cao-etal-2023-assessing,tao2024cultural}, 
and interaction with users \cite{shen2024towards,kirk2024prism}.
Moreover, recent theoretical frameworks \cite{uma2021learning,frenda2024perspectivist} underline the need to take into account various \textit{perspectives} about linguistic and pragmatic phenomena.
Especially the detection of subjective phenomena (i.e., toxic language) proved to be affected by different perceptions that reflect annotators' backgrounds, beliefs, values, and identities \cite{sap-etal-2022-annotators,fleisig-etal-2023-majority}. 
Therefore, a content moderation system should be representative of these different opinions, especially if these opinions come from segments of the population that are actually targets of attacks online \cite{kalluri2020don}. 
Focusing on toxic language detection, datasets like SBIC and CREhate with multiple annotations and information about annotators have been proposed and 
proved to be useful for investigating biases (as we have done in this work), building inclusive \cite{casola2023confidence} and personalized models \cite{kocon2021offensive}, and providing informative explanations about models' decisions \cite{mastromattei-basile-zanzotto:2022:NLPerspectives}.



\paragraph{Confidence and multiple annotations.}
The most common method to estimate confidence in models is the logit-based method that assesses their uncertainty using token-level probabilities employable to LLMs \cite{geng-etal-2024-survey} and other models \cite{wu2021logit}.  
In \citet{frenda2023epic}, softmax-based measure of uncertainty
has been employed to analyze the level of confidence of models trained on the annotations of specific segments of the population 
compared with a model trained on majority voting decision, showing that the formers
tend to make a decision with less uncertainty than the standard model.
Similar results are reported by \citet{anand-etal-2024-dont}, where the use of Multi-Ground Truths models, trained on instance-annotator label pairs, improved confidence for samples characterized by substantial annotation disagreements. In this last work, the confidence is computed as the mean class probability for each data's gold label across the epochs.

\paragraph{Conformal Prediction in NLP.}
Differently from previous works,
we rely on conformal prediction \cite{10.1561/2200000101}.
This framework
offers a systematic way to account for 
model
uncertainty, providing confidence measures that can inform decision-making, improve model interpretability, and mitigate biases in automated language processing systems.
One of its
key strengths 
is its ability to maintain robustness under distribution shifts, a property that has led to its widespread adoption in fields such as time series analysis \cite{shafer2008tutorial, papadopoulos2002inductive}. In many real-world applications, models are trained on datasets that may not fully capture the variability and complexity of the data they encounter in deployment. This discrepancy between training and deployment distributions—commonly referred to as distribution shift—can severely impact model performance and reliability. Conformal prediction helps mitigate this issue by providing statistically rigorous uncertainty estimates that remain valid even when the underlying data distribution changes \cite{vovk2005algorithmic}. This adaptability makes it particularly useful in settings where data evolves over time or where collecting perfectly representative training samples is infeasible, like in our case.
Following  
its success in domains like time series forecasting, 
conformal prediction has been employed recently in 
NLP \cite{campos-etal-2024-conformal}. 
\citet{villate2024collaborative}, for instance, used conformal prediction in a framework of content moderation based on model uncertainty in predicting toxicity and disagreement among annotators. Differently from our work, the authors do not focus on highlighting demographic-based biases in models to detect hate speech.

\section{Methodological Framework} \label{sec:framework}
Our methodology relies on conformal prediction, a statistical framework used to quantify the reliability of model predictions by assessing conformity, or how well individual predictions align with a set of labels \cite{angelopoulos2021gentle}. We leverage this theoretical framework to design two metrics for the analysis of bias in pre-trained models with the specific aim to measure their uncertainty against four socio-demographic groups based on the intersection of gender and ethnicity: 
white men, white women, non-white men, and non-white women.

\subsection{Uncertainty Divergence} \label{ss:uncertainty}
We utilized the \textit{Brier Score} as a core component to implement a conformal prediction framework to compare the average uncertainty of a model against a given annotator and the gold standard label obtained through majority vote. 


For a single annotated text, and a set of possible labels, $\mathcal{Y}$, the \textit{Brier Score} \( b(t_k) \) for text \( t_k \) can be written as
\begin{equation}
    b(t_k,\mathcal{Y}) = \frac{1}{|\mathcal{Y}|} \sum_{y \in \mathcal{Y}} \left( o_y(t_k) - p_M(y \mid t_k) \right)^2
    \label{eq:brier}
\end{equation}

where:
\begin{itemize}
    \item \( o_y(t_k) \) is the binary indicator (1 if the true label is \( y \), else 0).
    
    \item \( p_M(y \mid t_k) \) is the model-predicted probability for label \( y \).
\end{itemize}

The \textit{Brier Score} is directly used as a single conformity score to quantify the alignment of model predictions with observed outcomes. A lower score indicates better conformity, reflecting predictions that are less uncertain and better calibrated.

\paragraph{Conformity delta.} Since annotations often reflect the individual perspectives of annotators, beyond aggregated labels, we quantified prediction uncertainty by introducing the concept of the \textit{Conformity Delta} ($\Delta$). This measures the variability in the model's confidence when predictions are compared across individual and aggregated labels, providing deeper insights into uncertainty and reliability.

Let \( a \in \mathcal{A}= \{a_1,.., a_m\}\) be a single annotator, \(\mathcal{T} = \{t_1, \dots, t_n\}\) denote the set of annotated texts, and \(M\) represent an automatic annotation model that outputs a probability distribution over labels \(\mathcal{Y}\).  

For a text \(t_k \in \mathcal{T}\), let the label provided by the annotator \(a_i\) be \(y_{a}\in \mathcal{Y}\), and given the ground-truth label $y_{A}$ for the text \(t_k\), obtained as the majority score among a specific subset of annotators $A \subset \mathcal{A}$ (e.g., the annotators belonging to a particular demographic group), the uncertainty \(\delta(t_k)\) for the text \(t_k\) and the annotator \(a_i\) is defined as
\begin{equation}
    \delta_{a_i}(t_k) = b(t_k, y_{a_i}) - b(t_k, y_{A}),
    \label{eq:delta_sing}
\end{equation}

This $\delta_{a_i}(t_k)$ measures the variability in model confidence when predictions are evaluated against individual versus aggregated labels. A high value indicates significant disagreement or variability, often highlighting areas where annotators may have diverging perspectives or where model predictions fail to achieve consistency across groups.

Let $\Delta_{A}$ and $\Delta$  be defined as.
\begin{equation}
    \Delta_{A} = \{\delta_{a_i}(t_k)\} \quad a_i \in A, \quad k = 1, \dots, n
    \label{eq:incertezza1}
\end{equation}

Finally, we can also consider the fully disaggregated $\Delta$, being the set of all disaggregated labels:
\begin{equation}
    \Delta_{\mathcal{A}} = \{\delta_{a_i}(t_k)\} \quad a_i \in \mathcal{A}, \quad k = 1, \dots, n
\end{equation}

\paragraph{Uncertainty divergence.} The combination of Brier Score and Conformity Delta enables a nuanced assessment of model performance. While the Brier Score captures the overall prediction quality, the Conformity Delta highlights cases where individual annotators diverge significantly from the consensus label. The specific delta sets $\Delta_{A}$ and $\Delta_\mathcal{A}$ offer complementary advantages: $\Delta_{A}$ allows for targeted analysis of specific demographic groups to identify group-specific biases, while $\Delta_\mathcal{A}$ provides a comprehensive view across all annotators to capture the full spectrum of individual variations.

This divergence may indicate areas where the model struggles to generalize or where ground-truth labels are inherently ambiguous. By identifying such discrepancies, this approach enhances interpretability and guides iterative model refinement. This is particularly important in tasks like abusive content moderation, where decisions can amplify societal biases if models are not carefully evaluated. Discrepancies in annotator labels, influenced by cultural or personal factors, can lead to biased training data. By systematically quantifying uncertainty through the Conformity Delta, we can identify areas of potential bias and ensure that AI systems operate transparently and equitably.

In order to compute a potential correlation between the four socio-demographic groups of annotators and their average conformity deltas, we introduce the \textit{Uncertainty Divergence}. For each group, we convert the obtained conformity deltas in $\Delta_A$ in a distribution with three categories: Conformity $\delta_{a_i}(t_k)$ $< 0$, Conformity $\delta_{a_i}(t_k)$ $= 0$, Conformity $\delta_{a_i}(t_k)$ $> 0$. We compute the Kullback-Leibler divergence \cite{van2014renyi}, which is defined as: 
\begin{equation}
    D_{KL}(P \| Q) = \sum_{i \in \{<0, 0, >0\}} P(i) \log \frac{P(i)}{Q(i)}
    \label{eq:kl_div}
\end{equation}

Where $P$ is the distribution of conformity $\Delta$ assigned to all the disaggregated labels in the corpus and $Q$ the group-based one. For each model, we report the general uncertainty, under the label ``total'', defined as the average of the single uncertainties across the whole dataset. This enables measuring the general uncertainty of benchmarked models and specific uncertainties against socio-demographic groups.

\subsection{Demographic Divergence.} \label{ss:demographic}
To computationally represent annotators, we leveraged the concept of uncertainty derived from the computation of Conformity Delta (Equation \ref{eq:incertezza1}). 
The uncertainty interval falls within the range \([-1, 1]\). Both ends of the scale indicate maximal disagreement, where the model strongly favors a label different from the one chosen by the annotator, albeit in opposite directions. A value of \(0\) indicates perfect agreement, where the model aligns with the annotator’s label.

Each annotator \(a \in \mathcal{A}\) is finally represented by a 40-dimensional vector \(\mathbf{v}_{a} \in \mathbb{R}^{40}\), where each element \(\mathbf{v}_{a}[j]\) corresponds to the frequency of uncertainty values \(\delta\) for the texts annotated by \(a\) that fall within the \(j\)-th bin $\text{Bin}_j$. The number of bins was selected based on empirical experimentation: we varied the discretization granularity from 10 to 100 bins and observed the effect on clustering with KMeans. While the inertia naturally increases with higher-dimensional vectors due to geometric dispersion, the incremental gain in resolution starts to plateau around 40 bins, indicating that further increasing the number of bins does not substantially improve the discriminative power of the annotator representation.

Given that, we can define the value of the \(j\)-th element of the vector \(\mathbf{v}_{a}\) for the model $M$ as
\begin{equation}
 \mathbf{v^M}_{a_i}[j] = \frac{1}{|\mathcal{T}_{a_i}|} \sum_{t_k \in \mathcal{T}_{a_i}} \mathbb{I}(\Delta_{a_i}(t_k) \in \text{Bin}_j), 
 \label{eq:user_representation}
\end{equation}

where \(\mathcal{T}_{a_i} \subseteq \mathcal{T}\) is the set of texts annotated by \(a_i\), $\mathbb{I}(\cdot)$ is the indicator function, which equals \(1\) if the condition is true and \(0\) otherwise, and $\text{Bin}_j$ in one of the 40 equally sized bins defined as: 

\begin{equation}
\resizebox{.87\hsize}{!}{$
    \text{Bin}_j = \left[-1 + \frac{(j-1) \cdot 2}{40}, -1 + \frac{j \cdot 2}{40}\right), \quad j = 1, \dots, 40.$}
    \label{eq:binning}   
\end{equation}

The resulting vector $\mathbf{v^M}_{a}$ effectively characterizes the annotator's judgment patterns relative to the model: high values in bins near 0 would indicate an annotator who frequently agrees with the model, high values in bins near -1 or 1 would indicate an annotator who often strongly disagrees with the model, and overall the specific distribution across all 40 bins creates a unique ``uncertainty fingerprint'' for each annotator.

This representation allows for subsequent clustering of annotators based on similarities in their uncertainty profiles, providing insights into distinct annotation behaviors and potential subgroups within the annotator population.

\paragraph{Demographic Divergence.} For a given model \(M\), we assess how much the demographic class distributions vary across the four clusters. To quantify this, we compute the \textit{Jensen-Shannon Divergence (JSD)} \cite{MENENDEZ1997307} (weighted on the cluster size) across the distributions of demographic classes in the clusters, treating the four cluster distributions as components of a mixture model. Let $\pi_1, \ldots, \pi_4$ be the weights of the clusters (where $||c_i||$ is the number of annotators inside the cluster $i$ and $\pi_i = ||c_i||/\sum_{j=1}^4 ||c_i||$), and the demographic probability distributions inside the clusters are $P_1, \ldots, P_4$, for each language model, $M$, we define:
        \begin{equation}
        \resizebox{.87\hsize}{!}{$
        {\rm JSD}_{\pi_1, \ldots, \pi_4}(P_1, \ldots, P_4) 
        = \sum_{i=1}^4 \pi_i D( P_i \parallel M )$}
        \label{eq:jensen}
        \end{equation}
        
If the clustering, solely based on annotators' uncertainty, does not show significant differences in demographic distributions across clusters, the model can be considered fair, as the uncertainty is not influenced by annotators' demographic characteristics.
\section{Experimental Setup} \label{sec:method}
In our experimental setting, we provide two studies that rely on the conformal prediction framework to assess the presence of bias in pre-trained models against four socio-demographic groups based on the intersection between gender and ethnicity: white men, white women, non-white men, and non-white women. The first study explores if the adoption of uncertainty can be a predictor of biases against vulnerable groups (RQ1) and leverages the \textit{Uncertainty Divergence} (Section \ref{ss:uncertainty}). The second study assesses models' fairness in user representation (RQ2) through the \textit{Demographic Divergence} metric (Section \ref{ss:demographic}). We analyze the uncertainty of $8$ fine-tuned LMs and $3$ LLMs in the classification of hate speech.  We adopt as a benchmark two disaggregated corpora annotated for hate speech, which include information about annotators, such as gender and ethnicity. 
Therefore, we are able to measure the uncertainty of each model against specific communities of people.\footnote{All the experiments have been run on a RTX 3070 TI with the Hugging Face library \textit{transformers}. We adopt the default setup of each model as it is available on Hugging Face.} 

\paragraph{Pre-trained Models.}
In order to account for different generations of models, we considered for our benchmark study a set of transformer-based language models, including both fine-tuned LMs and prompted LLMs in a zero-shot setting. 
For fine-tuned LMs, 
we based our selection on NLP community adoption metrics: we included all models with at least $1,000$ downloads on HuggingFace\footnote{\url{https://huggingface.co/}} during November 2024. This criterion yielded 8 language models for our experiments, representing a broad spectrum of training methodologies. As a result, we identified $8$ LMs for our experiment that have been trained with a wide range of approaches:

\begin{itemize}
    \item IMSyPP \cite{kralj2022handling}. A BERT-based model \cite{kenton2019bert} trained on a multilingual corpus of hate speech messages gathered from Youtube and Twitter with disaggregated annotations. 
    \item HateBert \cite{caselli2020hatebert}. A retrained version of BERT based on RAL-E: a dataset of posts from banned SubReddits. 
    \item Dynabench \cite{vidgen2021lftw}. A model trained on a dynamically annotated dataset in which messages have been annotated through a multi-step process.
    \item Twitter-Roberta-Base \cite{antypas-camacho-collados-2023-robust}. A BERT-based model trained on a composition of $13$ corpora annotated for hate speech, misogyny, and other correlated phenomena.
    \item Refugees\footnote{\url{henrystoll/hatespeech-refugees}}. A model developed as a collaboration between UNHCR, the UN Refugee Agency, and Copenhagen Business School.  
    \item DistilRoberta~\cite{badmatr11x2023distilroberta}.
    A fine-tuned version of RoBERTa on the badmatr11x dataset~\cite{badmatr11x2023dataset}
    \item Pysentimiento \cite{perez2021pysentimiento}. Trained on the HatEval dataset \cite{basile2019semeval}, the model is part of a multilingual toolkit developed for the detection of hate speech, sentiment analysis, emotion, and irony. 
    \item MuRIL \cite{das2022data}. This 
    is a fine-tuned version of MuRIL on English abusive speech dataset.
\end{itemize}

Furthermore, we selected $3$ open-source LLMs to replicate the experiment in a zero-shot setting.

\begin{itemize}
        \item Mistral \cite{Mistral} is one of the first European LLMs developed by a start-up led by former scientists of Facebook-AI. 
        \item Olmo \cite{Groeneveld2024OLMoAT} is the LLM of AllenAI and is characterized by the careful implementation of toxicity filtering strategies from the pretraining corpora. 
        \item Bloom \cite{bloom} is the outcome of a series of workshops that involved hundreds of NLP scientists.
    \end{itemize}


\paragraph{Corpora.}
For our experiment, we chose two existing disaggregated corpora annotated for hate speech detection: the Social Bias Inference Corpus (SBIC) and CREHate. The rationale for choosing these resources is twofold: \textit{i.} they represent two different generations of perspectivist datasets; \textit{ii.} they significantly vary in their size and average number of annotations per message. 

\texttt{SBIC} \cite{sap2020social} is the first disaggregated corpus for hate speech detection that includes information about annotators' gender and ethnicity. The dataset consists of $44,671$ messages collected from multiple sources of previously annotated corpora for the same phenomenon. SBIC is composed of $146,254$ annotations with an average of $3.2$ annotations per message. The number of individual labels diverging from the gold standard label represents the 4.9\%

\texttt{CREHate} \cite{lee-etal-2024-exploring-cross} is the latest disaggregated corpus for hate speech detection. CREHate is composed of $1,580$ messages from existing hate speech corpora that were re-annotated. The dataset includes $42,546$ annotations with an average of $26.9$ annotations per message. The number of individual labels diverging from the gold standard label represents the 9.7\%

Further description of both corpora and annotators is provided in Appendix~\ref{appendix:appendix2}.





\section{STUDY 1: Models Uncertainty towards Socio-Demographic Groups} \label{sec:study1}
In this study we adopt Uncertainty Divergence (Section \ref{ss:uncertainty}) to compare models' performance with their uncertainty against texts labeled by annotators belonging to four socio-demographic groups: white men ($w.m.$), white women ($w.f.$), non-white men ($\neg w.m.$), and non-white women ($\neg w.f.$). 

\begin{table*}[ht]
    \centering
    \resizebox{\textwidth}{!}{\footnotesize
    \begin{tabular}{c|c|cccc|c|cccc}
        \multicolumn{11}{c}{\textbf{F\textsubscript{1} score}} \\
        \multicolumn{1}{l}{} & \multicolumn{5}{|c|} {\textbf{SBIC}} & \multicolumn{5}{c}{\textbf{CREhate}}\\
        \textbf{model} & \textbf{total}  & \textbf{$w.m.$} & \textbf{$w.f.$}      & \textbf{$\neg w.m.$}                & \textbf{$\neg w.f.$} &
                         \textbf{total}  & \textbf{$w.m.$} & \textbf{$w.f.$}      & \textbf{$\neg w.m.$}                & \textbf{$\neg w.f.$}  \\
        \midrule
        IMSyPP         & 0.41     & +58\text{e}{-3} &  -17\text{e}{-3}     &  \cellcolor{pink!25} -4\text{e}{-2} & \cellcolor{green!25} +1\text{e}{-2} &
                        0.33  & \cellcolor{pink!25} -16\text{e}{-3}  & \cellcolor{green!25} +1\text{e}{-3} &  -7\text{e}{-4}  & +6\text{e}{-3} \\
         HateBert      & 0.51  & \cellcolor{pink!25} -31\text{e}{-4} & -23\text{e}{-4} & \cellcolor{green!25} 55\text{e}{-4} & \cellcolor{green!25} 55\text{e}{-4} &
                       0.49  & \cellcolor{pink!25} -2\text{e}{-3} & \cellcolor{green!25}+1\text{e}{-3} &  \cellcolor{green!25} +1\text{e}{-3} & +3\text{e}{-4} \\
         Dynabench     & 0.29  &  -5\text{e}{-3} & +5\text{e}{-3} & \cellcolor{pink!25} -26\text{e}{-3} & \cellcolor{green!25} +35\text{e}{-3}& 
                        0.34  & \cellcolor{pink!25} -4\text{e}{-4} & \cellcolor{green!25} +13\text{e}{-3} &  +8\text{e}{-3} & +5\text{e}{-3}\\
         Twitter-Roberta-Base & 0.31  & +1\text{e}{-3} & +8\text{e}{-3} & \cellcolor{pink!25} -19\text{e}{-3} & \cellcolor{green!25} +27\text{e}{-3}& 
                        0.37  & -3\text{e}{-3}& \cellcolor{pink!25} -1\text{e}{-2} & \cellcolor{green!25} +4\text{e}{-3} & +3\text{e}{-3}\\
         Refugees       & \textbf{0.57}  & -7\text{e}{-3} & +6\text{e}{-3} &  \cellcolor{green!25}+29\text{e}{-3} & \cellcolor{pink!25} -9\text{e}{-3}&
                        \textbf{0.55 } & \cellcolor{green!25} +4\text{e}{-3} & +2\text{e}{-3} & \cellcolor{pink!25} -38\text{e}{-4} &  -24\text{e}{-4}\\
         DistilRoberta  & 0.44  & -11\text{e}{-3} & +6\text{e}{-3} & \cellcolor{pink!25} -19\text{e}{-3} & \cellcolor{green!25} +1\text{e}{-2}&
                          0.44  & \cellcolor{pink!25}-1\text{e}{-2} &\cellcolor{green!25} +9\text{e}{-3} & +6\text{e}{-4} & -4\text{e}{-4}\\
         Pysentimiento  & 0.35  & 8\text{e}{-3}  & 2\text{e}{-3} & \cellcolor{pink!25} -58\text{e}{-4} & \cellcolor{green!25}+26\text{e}{-3}&
                         0.33  & \cellcolor{pink!25} -19\text{e}{-3}  & \cellcolor{green!25} +13\text{e}{-3} & \cellcolor{green!25} +13\text{e}{-3} &  +9\text{e}{-3}\\
         MuRIL          & 0.36 & 10\text{e}{-3} & +12\text{e}{-6}  & \cellcolor{pink!25}-55\text{e}{-4} & \cellcolor{green!25}12\text{e}{-3}&
                        0.31  & \cellcolor{pink!25} -17\text{e}{-3} & \cellcolor{green!25}+12\text{e}{-3} & +6\text{e}{-4} & +5\text{e}{-4}\\
         \hline
         \hline
        Olmo-7B         &    0.48  &  -68\text{e}{-4} & +35\text{e}{-5} & \cellcolor{green!25} +51\text{e}{-3} & \cellcolor{pink!25}-12\text{e}{-3}& 
                       0.55  & \cellcolor{green!25} 12\text{e}{-3} & -44\text{e}{-4} & \cellcolor{pink!25}-74\text{e}{-4} & -21\text{e}{-4}\\
        Bloom-7B       & 0.48  & \cellcolor{pink!25} -30\text{e}{-4} & -14\text{e}{-4} & \cellcolor{green!25}+19\text{e}{-3} & +16\text{e}{-3}& 
                       0.49  & \cellcolor{green!25} +22\text{e}{-4} & \cellcolor{pink!25}-65\text{e}{-4} & +6\text{e}{-4} & +11\text{e}{-4}\\
        Mistral-7B     & 0.49 & \cellcolor{pink!25} -23\text{e}{-4} & +31\text{e}{-4} & -21\text{e}{-4} & \cellcolor{green!25} +53\text{e}{-4}&
                       0.50  & \cellcolor{pink!25} -67\text{e}{-4} & \cellcolor{green!25}+75\text{e}{-4} & +24\text{e}{-4} & -40\text{e}{-4}\\
    \end{tabular}
    }
    \caption{Delta $F_1$ score for SBIC and CREhate obtained with each model against the total list of disaggregated labels and the $F_1$ score against labels of a specific group: white men ($w.m.$); white women ($w.f.$); non-white men ($\neg w.m.$), non-white women ($\neg w.f.$). Models that encode optimally the perspective of a group are highlighted in green. Models that predict worse on the group than on majority voting are highlighted in red.}
    \label{tab:conformities_fscore}
\end{table*}

We compute the $F_1$ score obtained with each model against the total list of disaggregated labels and the $F_1$ score against labels of a specific group. This enables ranking the general performance of each model and observing differences between groups. Table \ref{tab:conformities_fscore} shows results of this analysis. As can be observed, the Refugees model obtains the best $F_1$ score on both SBIC ($0.57$) and CREHate datasets ($0.55$, shared with Olmo-7B). The analysis broken down by groups shows a distinction on the gender axis. In $15$ cases out of $22$, all models are better at predicting labels annotated by women. In SBIC, models perform a higher $F_1$ annotation of non-white women; in CREHate white women. A second pattern is about the performance of LLMs. In both cases, a pattern based on race emerges. Their predictions are better on non-white people in SBIC, while the opposite is observable for CREHate.

The results in Table \ref{tab:conformities_delta} show that LLM predictions are more prone to uncertainty: $2$ out $3$ obtain the highest average conformity $\Delta$. Observing the divergence of each group, it is possible to identify a systematic lower uncertainty in the classification of men's labels: white in SBIC, non-white in CREHate.

\begin{table*}[ht!]
    \centering
    \resizebox{\textwidth}{!}{\footnotesize
    \begin{tabular}{c|c|cccc|c|cccc}
        \multicolumn{11}{c}{\textbf{Uncertainty Divergence}} \\
        \multicolumn{1}{l}{} & \multicolumn{5}{|c|} {\textbf{SBIC}} & \multicolumn{5}{c}{\textbf{CREhate}}\\
        \textbf{model} & \textbf{total} &\textbf{$w.m.$} & \textbf{$w.f.$} & \textbf{$\neg w.m.$} & \textbf{$\neg w.f.$}& \textbf{total} & \textbf{$w.m.$} & \textbf{$w.f.$} & \textbf{$\neg w.m.$} & \textbf{$\neg w.f.$}\\
        \midrule
         IMSyPP &  -3\text{e}{-4} &\cellcolor{green!25} 55\text{e}{-5}  & 88\text{e}{-5} & \cellcolor{pink!25} 10\text{e}{-3}  &  66\text{e}{-4} & 
         -19\text{e}{-4} & \cellcolor{pink!25} 19\text{e}{-4}  &12\text{e}{-4} & \cellcolor{green!25} 29\text{e}{-5}  & 16\text{e}{-4} \\
         
         HateBert  & -12\text{e}{-4}& \cellcolor{green!25} 49\text{e}{-6} & 67\text{e}{-5} & 34\text{e}{-4} & \cellcolor{pink!25} 45\text{e}{-4} & -5\text{e}{-4}& 
         83\text{e}{-5} & 68\text{e}{-5} &  \cellcolor{green!25} 40\text{e}{-5} & \cellcolor{pink!25} 16\text{e}{-4} \\
         
         Dynabench & -7\text{e}{-4}& \cellcolor{green!25} 59\text{e}{-6} & 76\text{e}{-5} & \cellcolor{pink!25} 36\text{e}{-4} & 45\text{e}{-5}& -9\text{e}{-4}&
         84\text{e}{-5} &88\text{e}{-5} & \cellcolor{green!25} 32\text{e}{-5} & \cellcolor{pink!25} 18\text{e}{-4}\\
         
         Twitter-Roberta-Base & -15\text{e}{-4}& 15\text{e}{-5} & \cellcolor{green!25} 10\text{e}{-5} & \cellcolor{pink!25} 45\text{e}{-4} & \cellcolor{pink!25} 45\text{e}{-4}& 13\text{e}{-4}&
          19\text{e}{-5} & \cellcolor{green!25} 13\text{e}{-5} & \cellcolor{pink!25} 10\text{e}{-3} & 66\text{e}{-4}\\
         
         Refugees &13\text{e}{-4}&
                          19\text{e}{-5} & \cellcolor{green!25} 13\text{e}{-5} & \cellcolor{pink!25} 10\text{e}{-3} & 66\text{e}{-4}& 4\text{e}{-4}&
         11\text{e}{-4} & 55\text{e}{-5} & \cellcolor{green!25} 39\text{e}{-5} &  \cellcolor{pink!25} 15\text{e}{-4}\\
         
         DistilRoberta & -3\text{e}{-4}&\cellcolor{green!25} 43\text{e}{-6} & 70\text{e}{-5} & 35\text{e}{-4} & \cellcolor{pink!25} 46\text{e}{-4}&
                     -12\text{e}{-4}&94\text{e}{-5} & 72\text{e}{-5} & \cellcolor{green!25} 29\text{e}{-5} & \cellcolor{pink!25} 16\text{e}{-4}\\
                     
         Pysentimiento & 14\text{e}{-4}& \cellcolor{green!25} 91\text{e}{-5}  & 20\text{e}{-4} & \cellcolor{pink!25}58\text{e}{-4} & 47\text{e}{-4}&
                         -23\text{e}{-4}&\cellcolor{pink!25} 23\text{e}{-4}  & 12\text{e}{-4} & \cellcolor{green!25} 30\text{e}{-5} &  16\text{e}{-4}\\
                         
         MuRIL & -71\text{e}{-4}&\cellcolor{green!25} 10\text{e}{-4} & 20\text{e}{-4}  & \cellcolor{pink!25} 59\text{e}{-4} & 49\text{e}{-4}& 
                             19\text{e}{-5}&\cellcolor{pink!25} 23\text{e}{-4} & 12\text{e}{-4} & \cellcolor{green!25} 31\text{e}{-5} & 16\text{e}{-4}\\
         \hline
         \hline
         Olmo-7B  & 16\text{e}{-4}& \cellcolor{green!25} 54\text{e}{-5} & 18\text{e}{-4} & \cellcolor{pink!25} 10\text{e}{-3} &  30\text{e}{-4}& 
                      56\text{e}{-5}& 13\text{e}{-4} & 6\text{e}{-4} & \cellcolor{green!25} 4\text{e}{-4} & \cellcolor{pink!25} 15\text{e}{-4}\\
         Bloom-7B  & 16\text{e}{-4}&\cellcolor{green!25} 15\text{e}{-5} & 10\text{e}{-4} &  39\text{e}{-4} & \cellcolor{pink!25} 42\text{e}{-4}&
                     32\text{e}{-4}& 14\text{e}{-4} & 9\text{e}{-4} & \cellcolor{green!25} 3\text{e}{-4} & \cellcolor{pink!25} 15\text{e}{-4}\\
         Mistral-7B  & -14\text{e}{-4}&\cellcolor{green!25} 84\text{e}{-6} & 10\text{e}{-4} &  39\text{e}{-4} & \cellcolor{pink!25} 42\text{e}{-4}& 
                    10\text{e}{-4}& 8\text{e}{-4} & 6\text{e}{-4} & \cellcolor{green!25} 3\text{e}{-4} & \cellcolor{pink!25} 15\text{e}{-4}\\
    \end{tabular}
    }
    \caption{The Uncertainty Divergence between the distribution of non-conformity scores assigned by models to all the individual annotations against the distributions of annotations grouped by the intersection of gender and ethnicity. The higher the score (red), the higher the divergence between the non-conformity of all the individual annotations and socio-demographic groups. The lower the score (green), the lower the divergence.}
    \label{tab:conformities_delta}
\end{table*}


The study shows that Uncertainty Divergence might be a reliable metric to investigate social biases emerging in classification. The metric does not correlate with the performance of models \textbf{[RQ1]}. Computing the T-Test \cite{kim2015t} between $F_1$ scores and conformity $\Delta$'s show that these two scores are not correlated both in SBIC ($p=0.14$) and CREHate ($p=0.11$). This suggests that models with a higher performance but a lower conformity might fail in understanding unseen messages outside corpora distribution. In this sense, the higher conformity $\Delta$ assigned to annotations provided by non-white people can be interpreted as \textbf{a predictor of a potential systematic misalignment between street-level algorithms decisions and women perception of hate speech}. Therefore, conformity might be not only used as a metric of fairness but as a guiding principle for selecting for content moderation models that are able to \textit{see through the lens} of vulnerable minorities.

\section{STUDY 2: User representation: Clustering annotators according to their uncertainty} \label{sec:study2}
\begin{figure}[ht]
    \centering
    \includegraphics[width=1\linewidth]{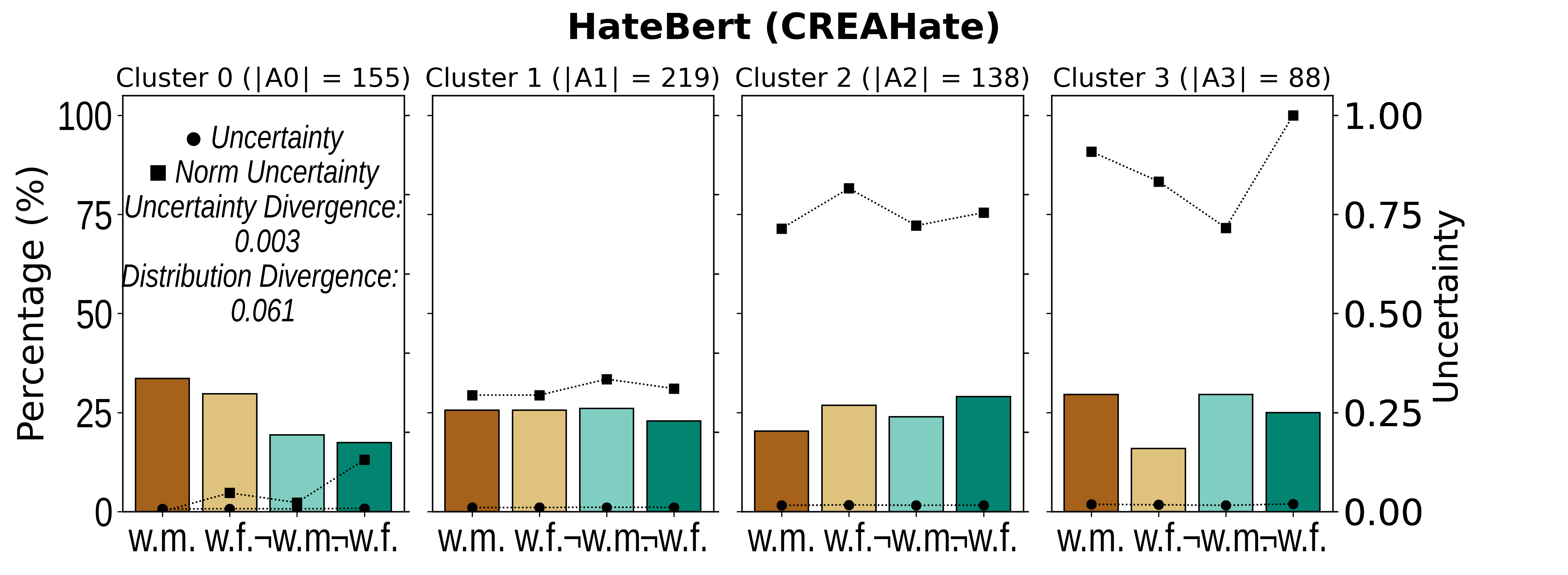}
    \caption{An example of annotator clusters based on the uncertainty of HateBert over the CREHate dataset.}
    \label{fig:clusters}
\end{figure}

This study leverages Demographic Divergence to represent users through models' uncertainty in text classification.     
The annotators were grouped into four distinct clusters for each model $M$ as described in Section \ref{ss:demographic}. The number of clusters was set to 4 to align with the examined socio-demographic groups ($w.m.$, $w.f.$, $\neg w.m.$, $\neg w.f.$).
Clustering was performed on the annotator vectors \(\mathbf{v^M}_{a}, \forall a \in A\), which represent the distribution of uncertainty for each annotator, using \(k\)-means. Figure \ref{fig:clusters} shows an example of results based on HateBert uncertainty scores over the CREHate dataset. Bar plots represent the demographic distribution of annotators in each cluster, and line plots represent the uncertainty related to each socio-demographic group. All the clusters are available in Appendix \ref{appendix:appendix1}.    

At this stage, for each of the 11 models considered, we have two clusterings (one for the CREHate dataset and one for the SBIC dataset), each consisting of 4 clusters. Our goal is twofold: on the one hand, to quantify the distribution of demographic characteristics of the annotators within the clusters, and on the other, to evaluate Uncertainty Divergence within individual clusters across the models. 

For a model $M$
and the four identified annotator clusters, we calculate the average uncertainty across demographic classes for each cluster by leveraging Uncertainty Divergence metrics.
If the average uncertainty varies across the different clusters, it indicates that there are groups of annotators, for which the model $M$ exhibits lower fairness compared to others.

\begin{figure}[ht]
    \centering
    \includegraphics[width=0.7\linewidth]{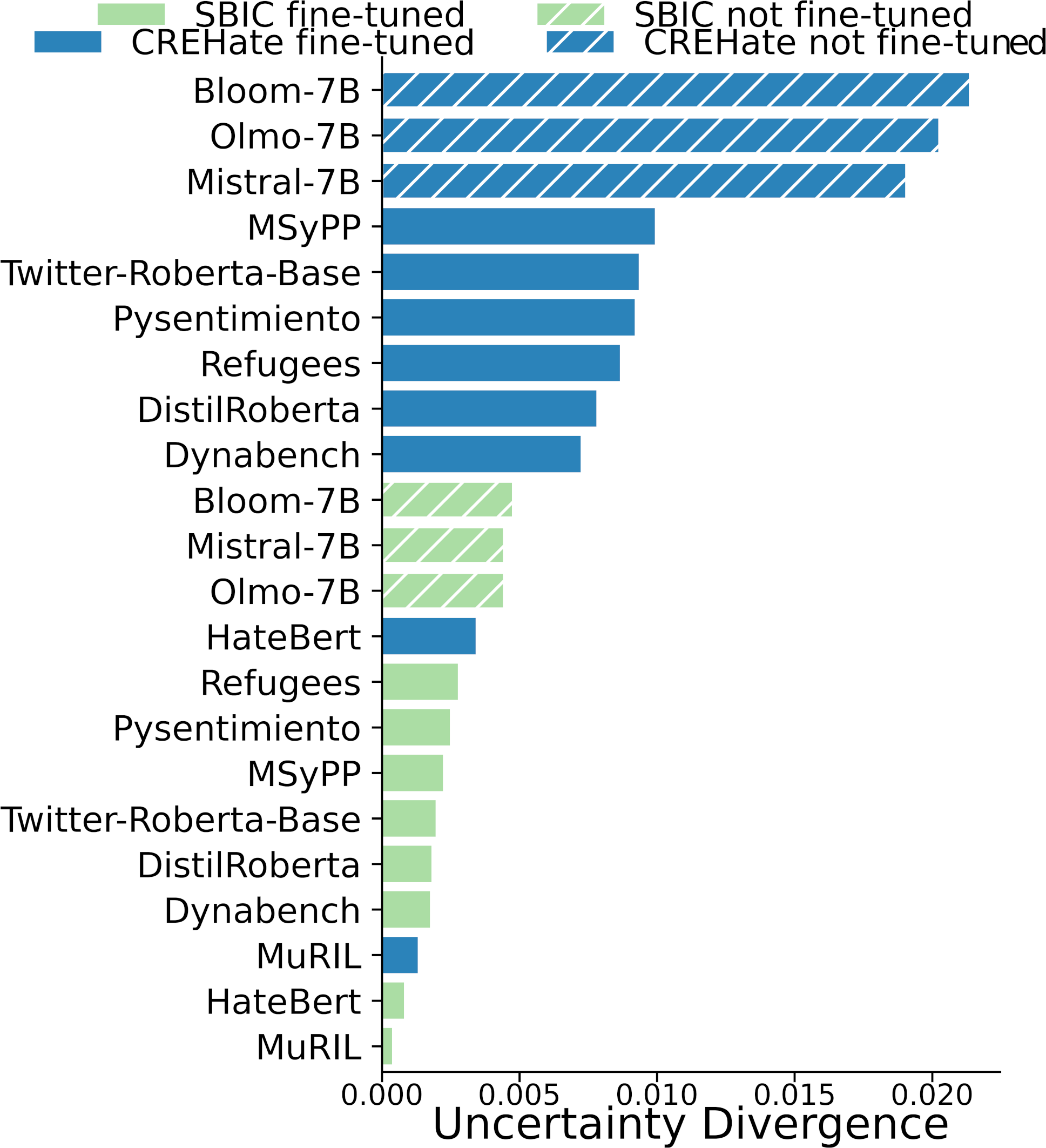}\\[1ex]
    \includegraphics[width=0.7\linewidth]{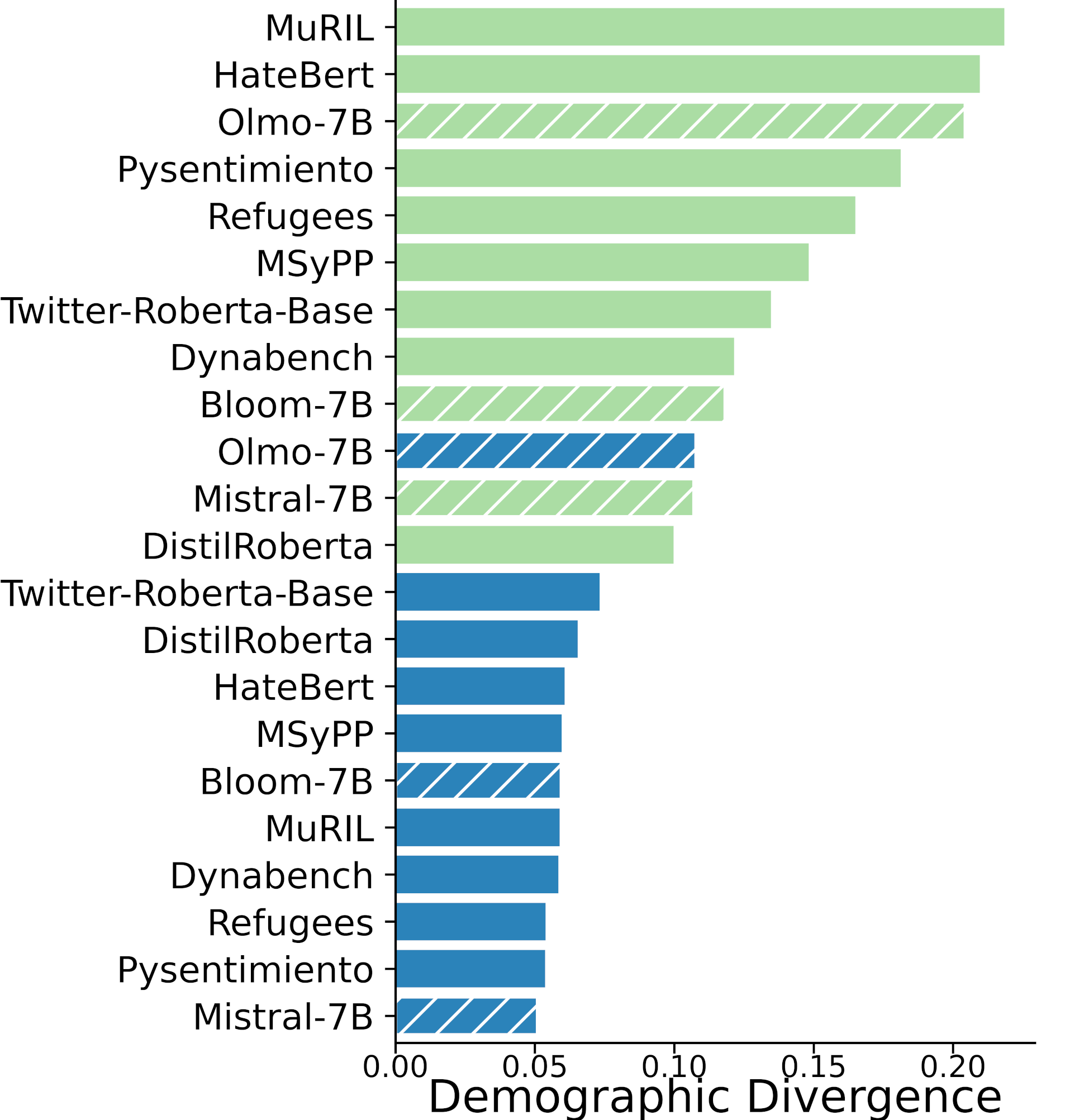}
    \caption{Ranking of the models based on Uncertainty Divergence and Demographic Divergence.}
    \label{fig:ranking}
\end{figure}

Figure~\ref{fig:ranking} illustrates the ranking of models based on Uncertainty Divergence and Demographic Divergence. 
%
%
Uncertainty Divergence and Demographic Divergence are specifically used to quantify the fairness of models \textbf{[RQ2]}.
By examining the figure, we can see that the LLMs exhibit higher Uncertainty Divergence for both datasets. Therefore, despite achieving 
optimal
results in terms of $F_1$, the higher uncertainty compared to the LMs indicates that there are groups of annotators for which LLMs exhibit lower fairness.

In contrast, Demographic Divergence helps us to understand whether this uncertainty is distributed equitably across the clusters, and hence, across the demographic classes. LLMs perform better in the ranking, particularly Mistral-7b. Consistent with Tables \ref{tab:conformities_fscore} and \ref{tab:conformities_delta}, this LLM maintains fairness across the considered dimensions (gender and ethnicity). On the other hand, Olmo-7b presents negative Demographic Divergence values, indicating higher uncertainty, which is not evenly distributed across the demographic classes.

We want to highlight the behavior of MuRIL, which exhibits the lowest uncertainty among all models for both datasets but presents the highest Demographic Divergence. In Appendix~\ref{appendix:appendix1}, Figure P shows that clusters with female or non-white female annotators exhibit higher uncertainty, and the distributions between clusters are significantly different.


\section{Discussion} \label{sec:findings}
Our research shows that uncertainty and the conformal prediction framework are \textbf{a powerful approach for the analysis of models' fairness in automatic content moderation}. Uncertainty is effective for the observation of the alignment of models' behavior with the sensibility of specific groups of annotators. Uncertainty is also a powerful way to understand how single annotators are represented and grouped, and to what extent models perpetuate social biases during this process.

From the benchmark analysis of pre-trained models, both general patterns and model-specific behaviors emerge. The vast majority of models show the lowest uncertainty in predicting contents annotated by men and the highest uncertainty for the prediction of contents annotated by non-white people. This implies that in the automatic content moderation settings, in which the model performance might suffer a high degradation due to temporal semantic shifts, the risk of misalignment between the model's predictions and human annotations is higher for non-white people. An explanation of such a pattern might be in the long-term impact of the pretraining process. Models trained on data that poorly represent non-white people learn perspectives of the world that are not easy to be removed. In this sense, the more nuanced results in terms of performance might be interpreted as an optimization over specific benchmark corpora, while \textbf{uncertainty could be the blueprint of pretraining biases}.

Despite the presence of common behaviors, choosing one pre-trained model over another matters. The average uncertainty of these models significantly varies without correlating with performance (Section \ref{sec:study1}), showing that the two ways of measuring them ($F_1$ score and Conformity $\Delta$) grasp different aspects of the same problem. In this sense \textbf{Mistral represents the best trade-off between performance and uncertainty}, despite the lack of information about which data has been used to train it \cite{Mistral}. However, it is not possible to draw conclusions from the comparison of fine-tuning and zero-shot approaches, since OLMO and Bloom exhibit a higher Demographic Divergence (Section \ref{sec:study2}) within their clusters: a predictor of bias against vulnerable groups to discrimination. 

Corpora themselves appear to be a factor in stressing the uncertainty of models and represent a significant limitation for fairness studies. Model predictions on SBIC systematically suffer higher Demographic Divergence (Appendix \ref{appendix:appendix1}) than predictions on CREHate. The very different composition of annotators (Section \ref{sec:method}) might play a role in this, as well as the different degrees of subjectivity emerging from raw annotations ($4.9\%$ \textit{vs} $9.7\%$), and the average number of annotations per message ($3.2$ \textit{vs} $26.9$). Furthermore, the inter-annotator agreement between annotators belonging to different demographic classes is generally higher within the same class, while agreement between different classes tends to vary (Appendix \ref{appendix:appendix_cohen}). Despite these differences, \textbf{the two corpora share the same limitation: binarism in the annotators' selection process}. Non-binary people were almost not involved in the annotation process, hindering most insightful analyses that go beyond the traditional intersection of race and gender. Resources developed for fairness should be more effective in representing marginalized and invisible groups of people. 

\section{Conclusion and Future Work} \label{sec:conclusion}
In this paper, we presented a novel approach to assess the fairness of models through their uncertainty. We introduced metrics for measuring the impact of uncertainty against socio-demographic groups.
In particular, we leveraged our unsupervised approach based on conformal prediction to benchmark $11$ street-level algorithms on SBIC and CREHate datasets: $8$ LMs fine-tuned for hate speech detection and $3$ LLMs instantiated through a prompt-based method. The results show that measuring models' uncertainty unfolds systematic and hidden biases against non-white people, which do not emerge from performance-based metrics, such as the $F_1$ score [\textbf{RQ1}].

Moreover, we generated vector representations of annotators based on uncertainty scores emerging from models' predictions and used them to cluster annotators. The socio-demographic composition of the resulting clusters significantly varies between models, which show different degrees of fairness against women and non-white people [\textbf{RQ2}].

Future work goes in two directions. We will test the impact of considering uncertainty during fine-tuning and active learning (e.g., through Reinforcement Learning approach) to reduce bias in model prediction. We will explore the transferability of our methodology on contiguous tasks to hate speech detection and to other perspectivist corpora.

Limitations
\section*{Limitations}
In this work, our approach has been tested on hate speech detection; however, to validate its generalizability, we will further employ it on the detection of other subjective phenomena (i.e., when a higher human label variation is a sign of diverse subjectivities). Additionally, we only choose a subset of models for our analysis. This might result in overlooking models that actually show different patterns in the representation of vulnerable groups than the ones emerging in our analysis. 
Finally, we focus in particular on dimensions of gender and ethnicity common to both datasets used as samples for proving our methodology. However, we are aware that a binary classification for gender and ethnicity is far from the real world and could raise discussion \cite{larson-2017-gender}. 
Moreover, considering other identity axes, it is possible that hidden forms of discrimination could emerge. Nevertheless, our approach can be used with multiple categories across various dimensions.

\section*{Ethical Issues}
Since this research relies on secondary data, there are no ethical issues related to the collection and annotation of texts. Research biases related to these previous studies may still have an impact on the representation of human annotators emerging from our results.

\section*{Acknowledgments}
The work of S. Frenda is supported by the EPSRC project “Equally Safe Online” (EP/W025493/1).

\bibliography{custom}

\appendix

\section{Corpora and annotators description}
\label{appendix:appendix2}

\begin{figure}[ht]
        \centering
        \includegraphics[width=1\linewidth]{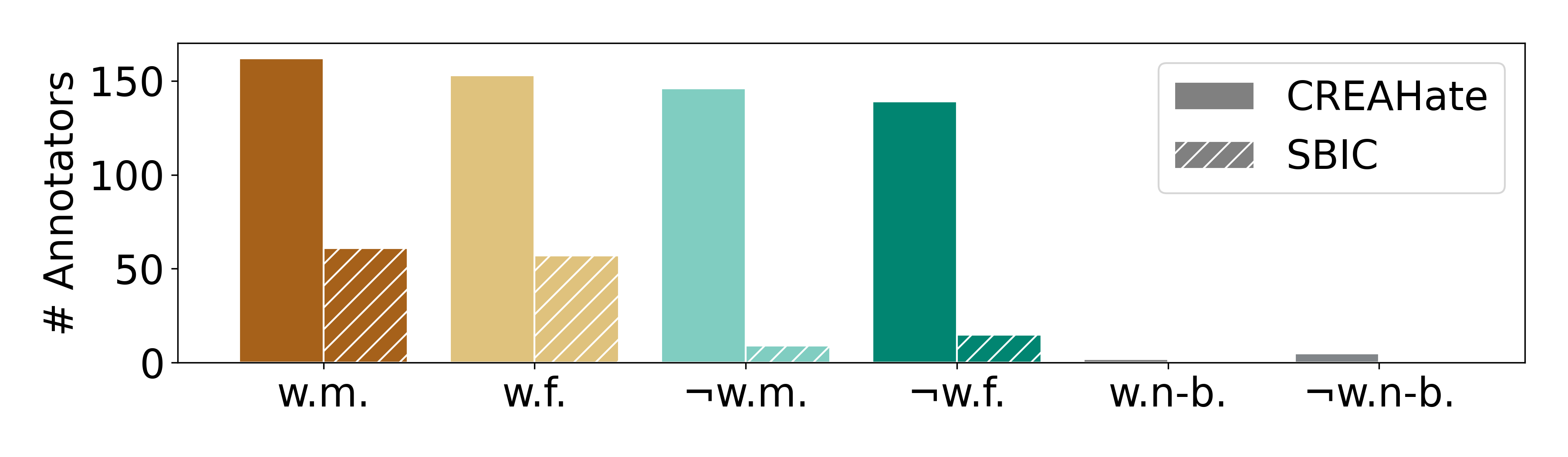} 
        \caption{Annotators' demographics on CREHate and SBIC. }
        \label{fig:annotatorDistribution}
    \end{figure}

Both datasets include information on the annotators' gender (male, female, non-binary) and ethnicity (White, Hispanic, Asian, etc.). Figure \ref{fig:annotatorDistribution} illustrates the distribution of the annotators' demographic classes. Non-binary individuals were excluded due to the low number of annotators in this gender category. For ethnicity, we grouped white and non-white annotators separately to achieve a more balanced distribution. Gender and ethnicity were then combined to form four distinct demographic classes: white man ($w.m$), white female ($w.f$), not-white male ($\neg w.m$), non-white female ($\neg w.f$) (Figure \ref{fig:annotatorDistribution}).

\begin{figure}[ht!]
        \centering
        \includegraphics[width=1\linewidth]{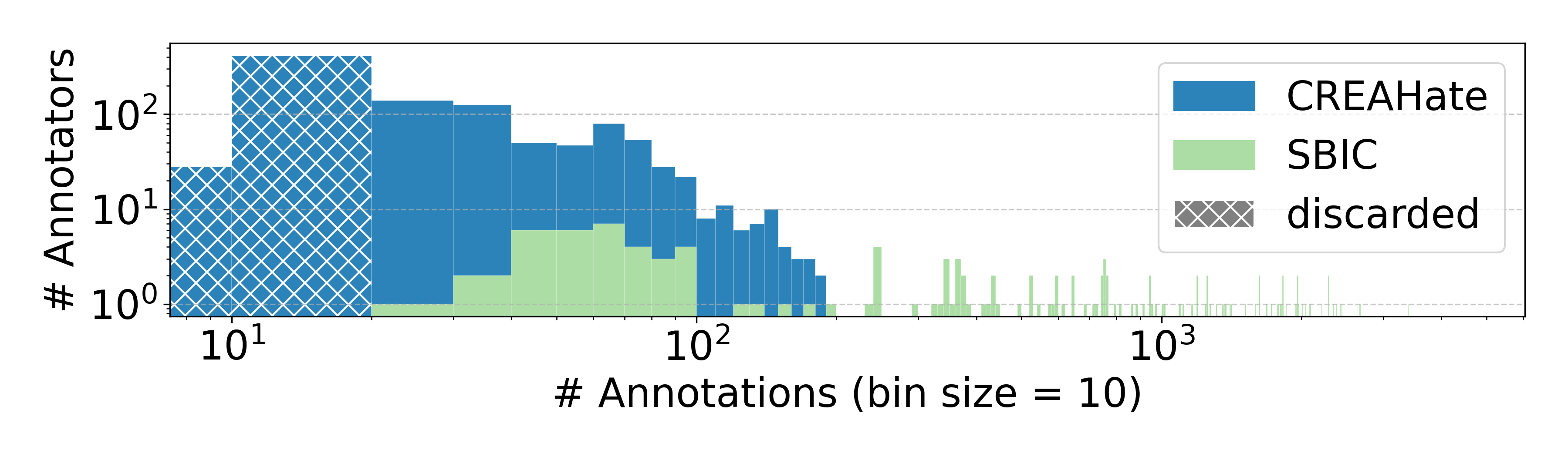}
        \caption{Distribution of the number of annotations per annotator.}
        \label{fig:annotationDistribution}
    \end{figure}

We excluded annotators who annotated fewer than 20 messages for two reasons. First, with too few messages, the uncertainty profile could be less reliable. Second, the threshold of 20 was chosen because the annotator with the fewest annotations in the SBIC dataset had annotated 24 messages. This approach ensured that, although the annotation distributions differed between datasets, they were made more comparable (Figure \ref{fig:annotationDistribution}).
    
\section{Uncertainty in LLMs and LMs}
\label{appendix:appendix1}

This appendix provides a fine-grained analysis of the findings discussed in Section~\ref{sec:study2}.
For each model,
we visualize the distribution of annotators' demographic classes alongside their corresponding uncertainty levels. Each row corresponds to a single model, with results from CREAHate on the left and SBIC on the right. Subplots [$a$–$o$] display the results for LMs, while [$s$–$v$] corresponds to LLMs. Notably, the SBIC dataset often results in very small annotator clusters. However, this does not affect the Uncertainty Divergence and Demographic Divergence metrics, as these are weighted by the number of annotators in each cluster. A key observation is that uncertainty—particularly when considering its normalized values across the four clusters—tends to be higher in clusters where women or non-white annotators are more prevalent (Figure \ref{fig:appendix}).

 \begin{figure*}[htbp]
        \subfloat[]{%
            \includegraphics[width=.48\linewidth]{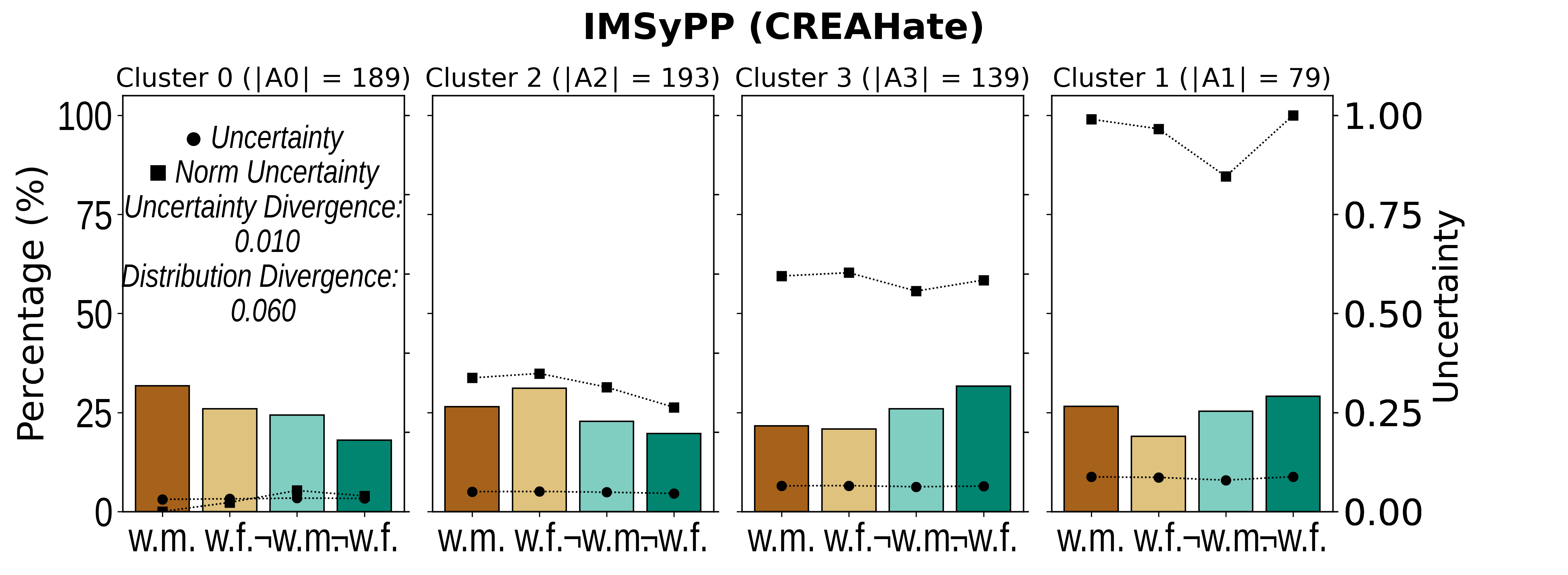}%
        }\hfill
        \subfloat[]{%
            \includegraphics[width=.48\linewidth]{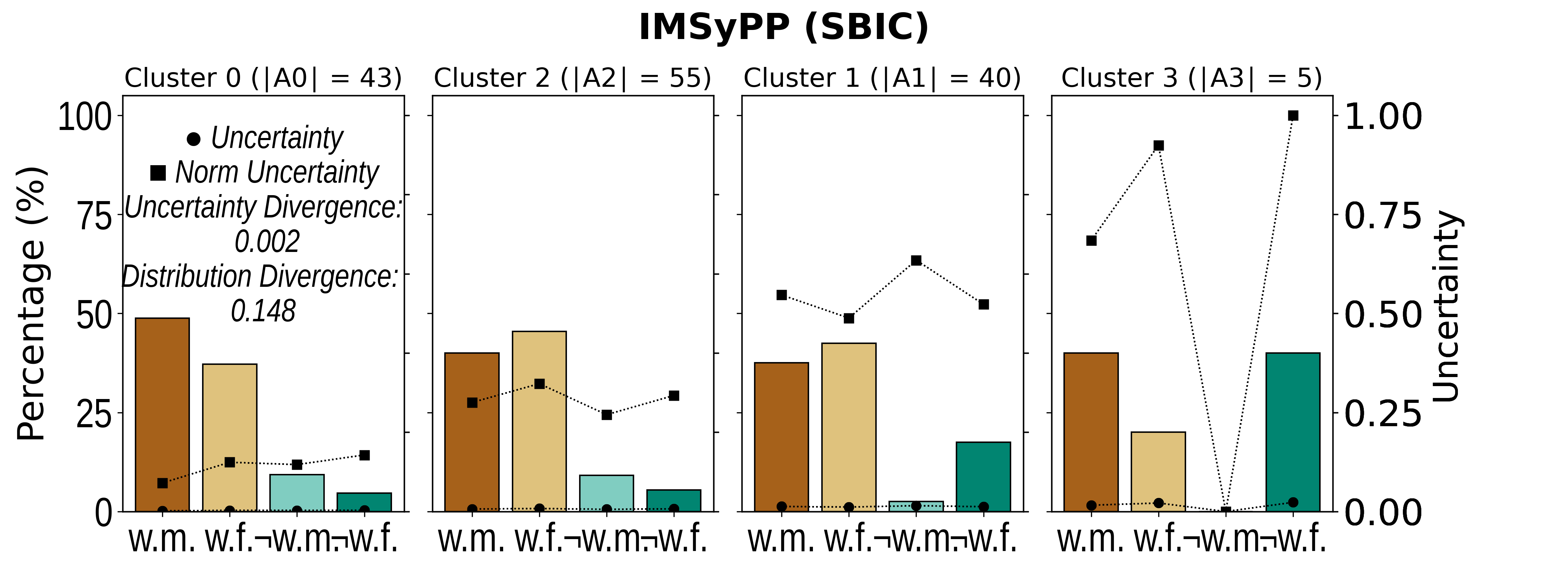}%
        }\\
        \subfloat[]{%
            \includegraphics[width=.48\linewidth]{Figure/modelli/crh_GroNLP-hateBERT_aggr.csv.png}%
        }\hfill
        \subfloat[]{%
            \includegraphics[width=.48\linewidth]{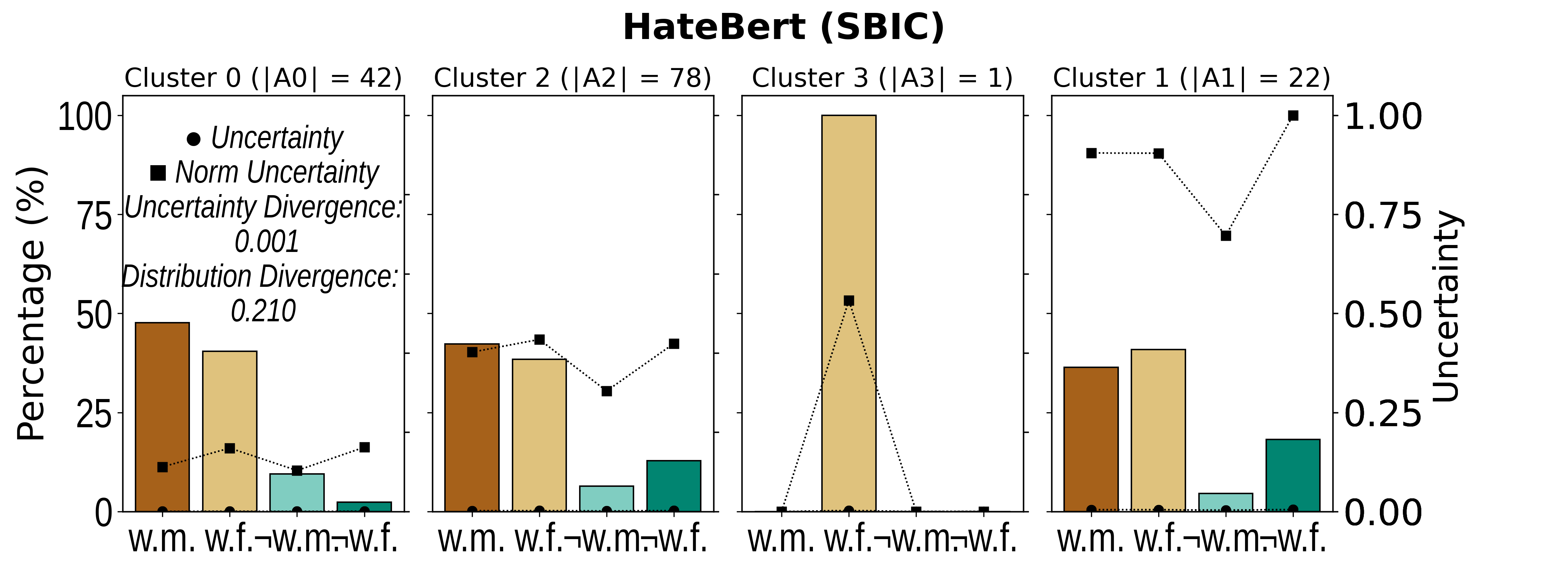}%
        }\\
        \subfloat[]{%
            \includegraphics[width=.48\linewidth]{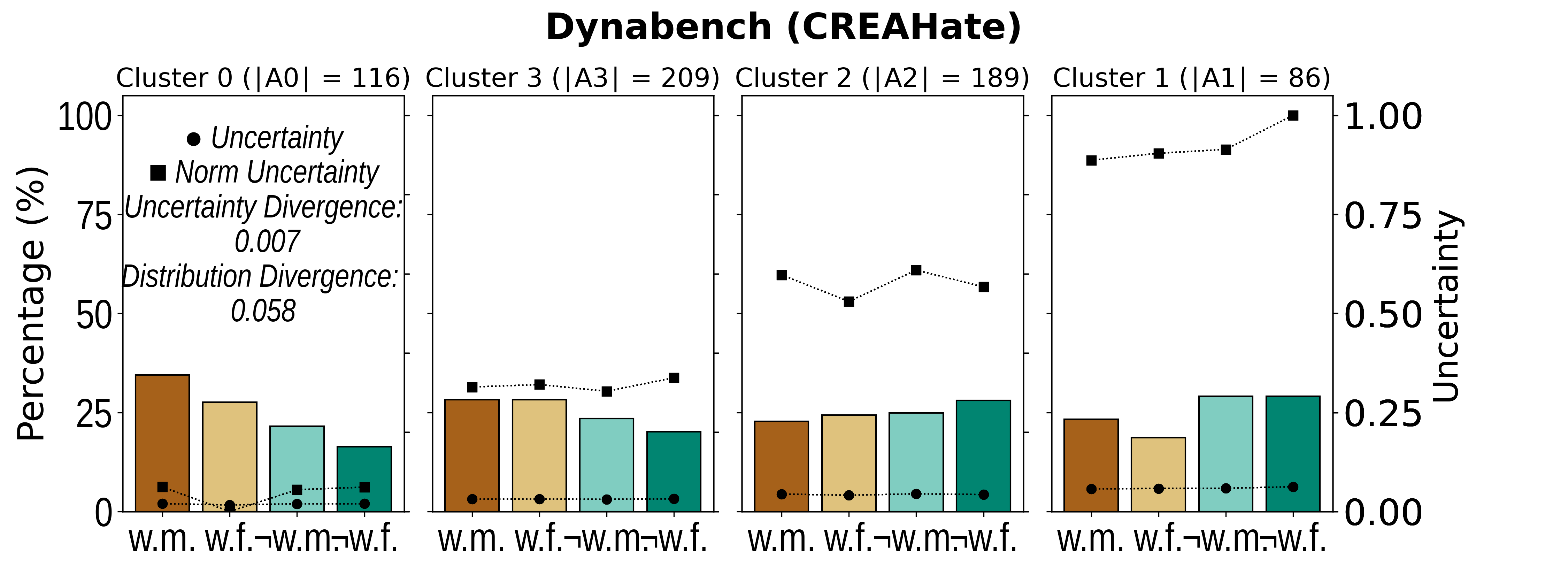}%
        }\hfill
        \subfloat[]{%
            \includegraphics[width=.48\linewidth]{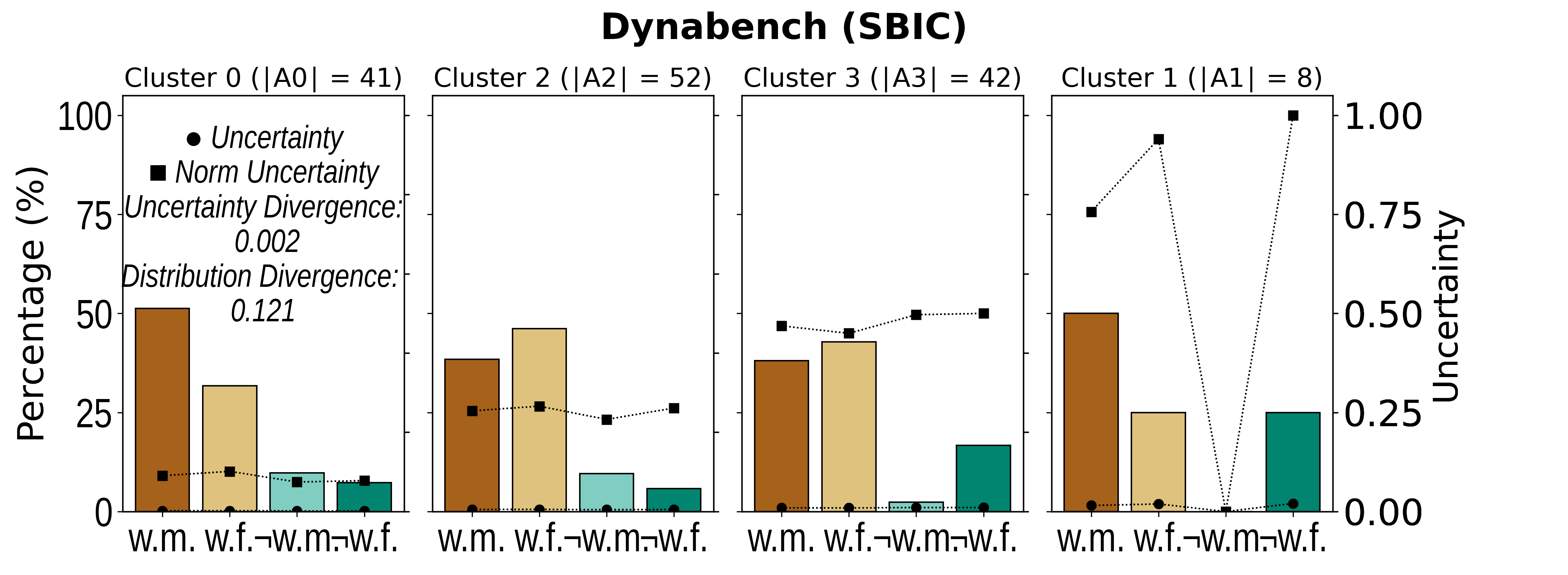}%
        }\\
        \subfloat[]{%
            \includegraphics[width=.48\linewidth]{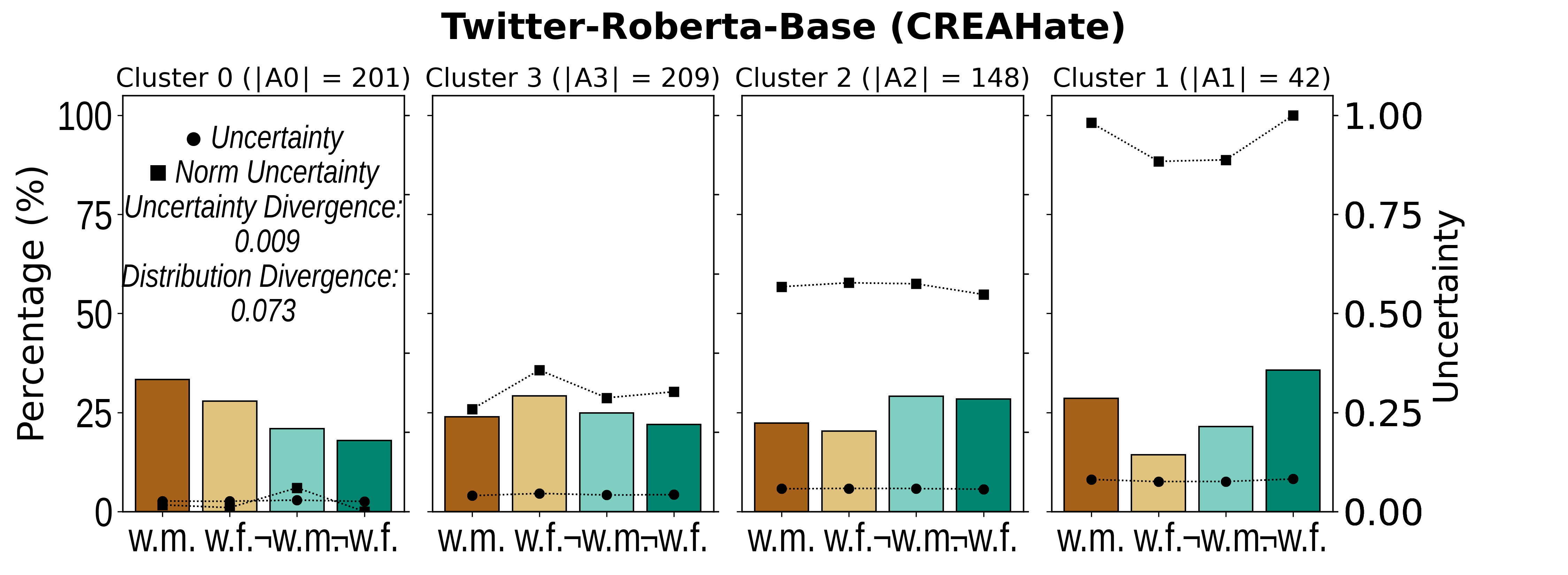}%
        }\hfill
        \subfloat[]{%
            \includegraphics[width=.48\linewidth]{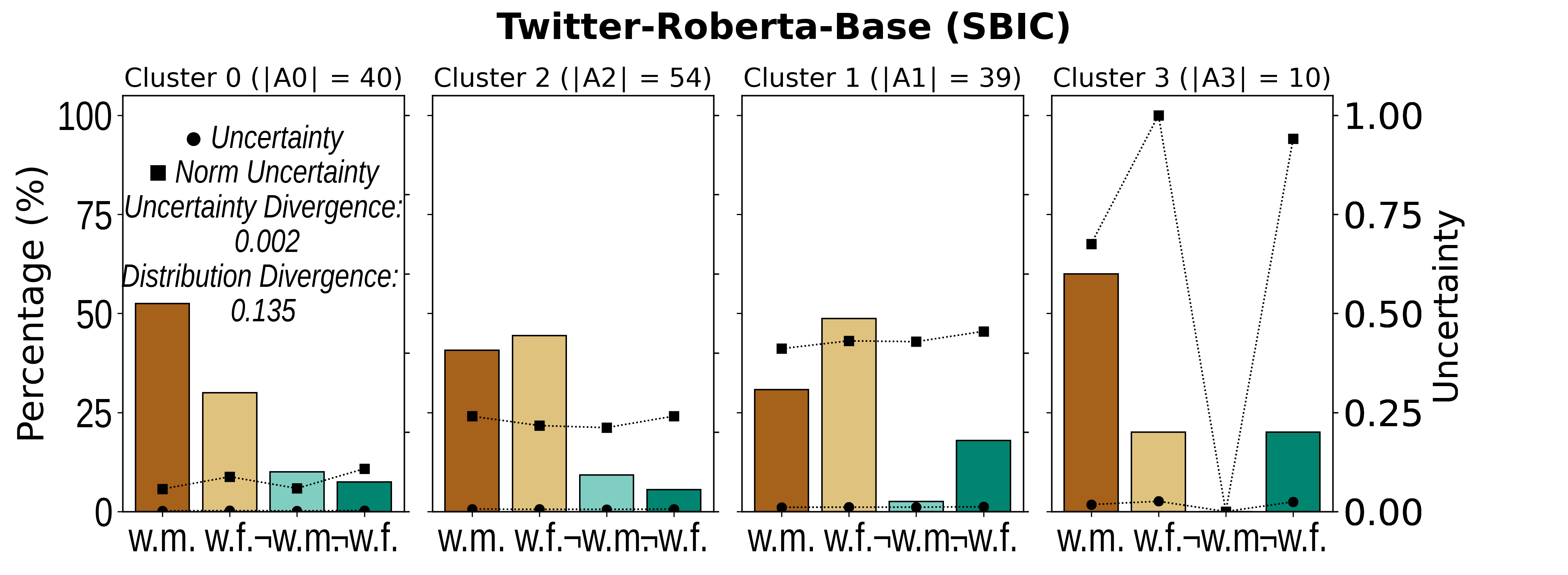}%
        }\\
        \subfloat[]{%
            \includegraphics[width=.48\linewidth]{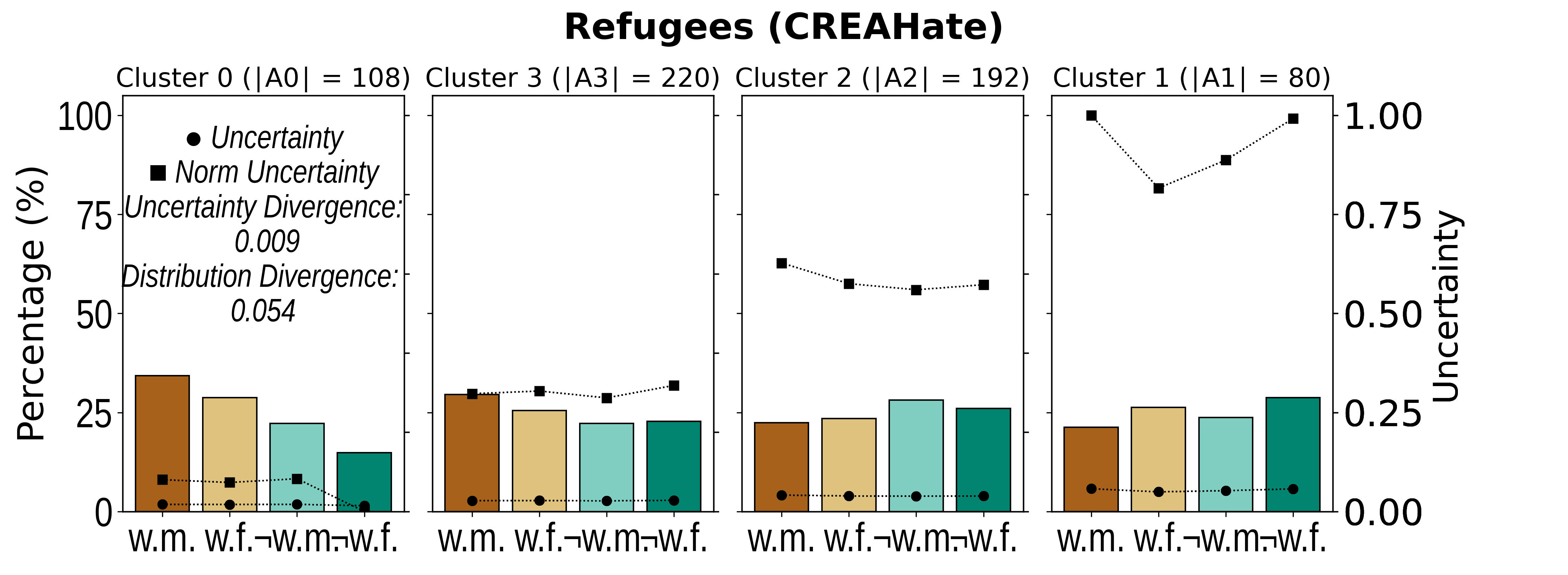}%
        }\hfill
        \subfloat[]{%
            \includegraphics[width=.48\linewidth]{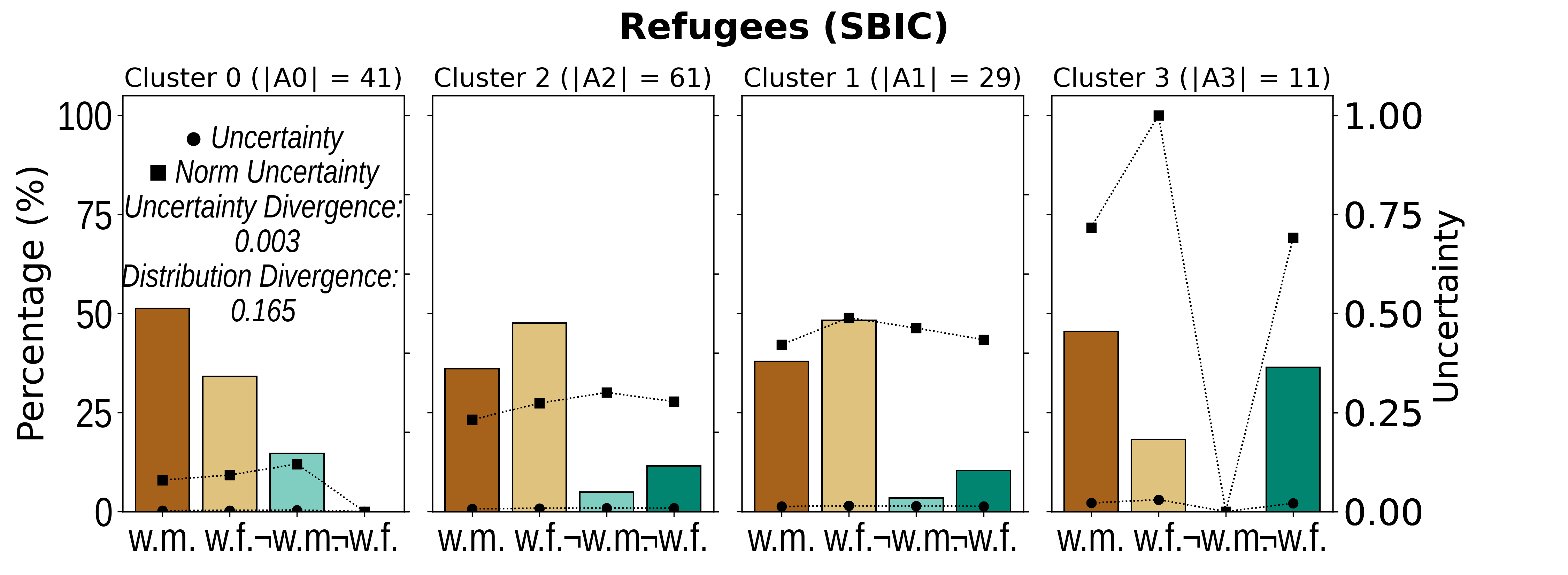}%
        }

    \end{figure*}

    \begin{figure*}[htbp]
        \ContinuedFloat
        \subfloat[]{%
            \includegraphics[width=.48\linewidth]{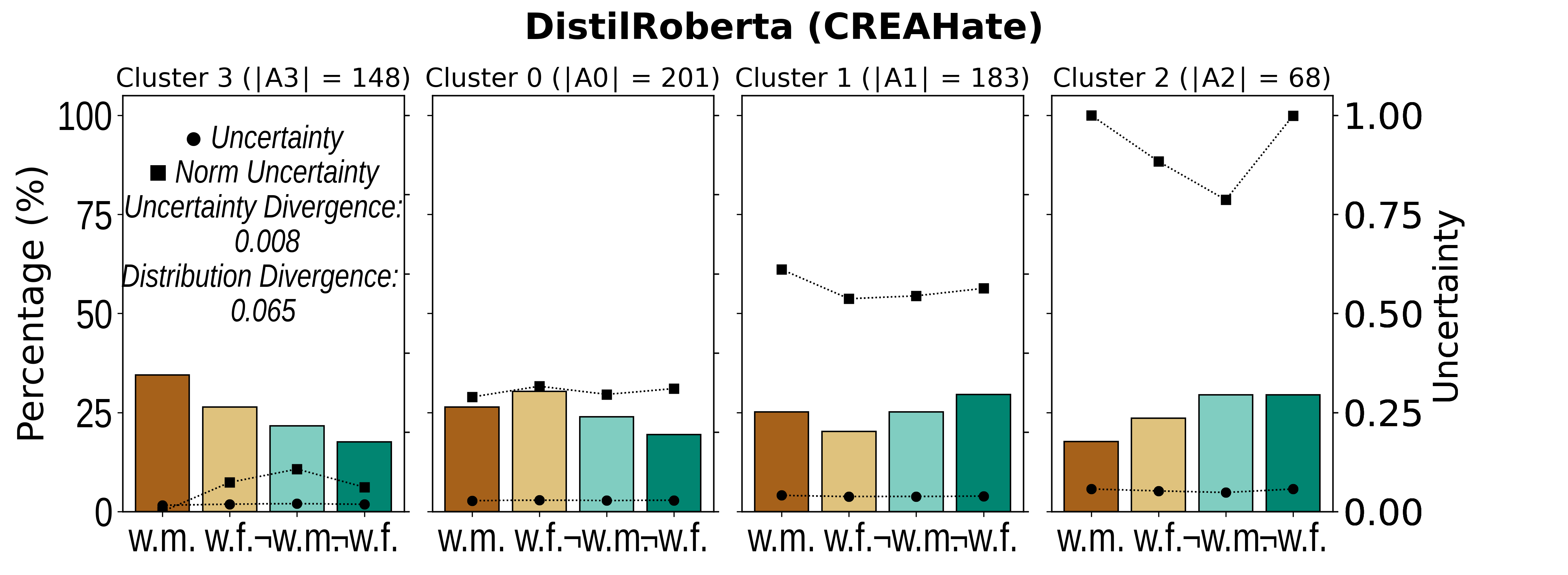}%
        }\hfill
        \subfloat[]{%
            \includegraphics[width=.48\linewidth]{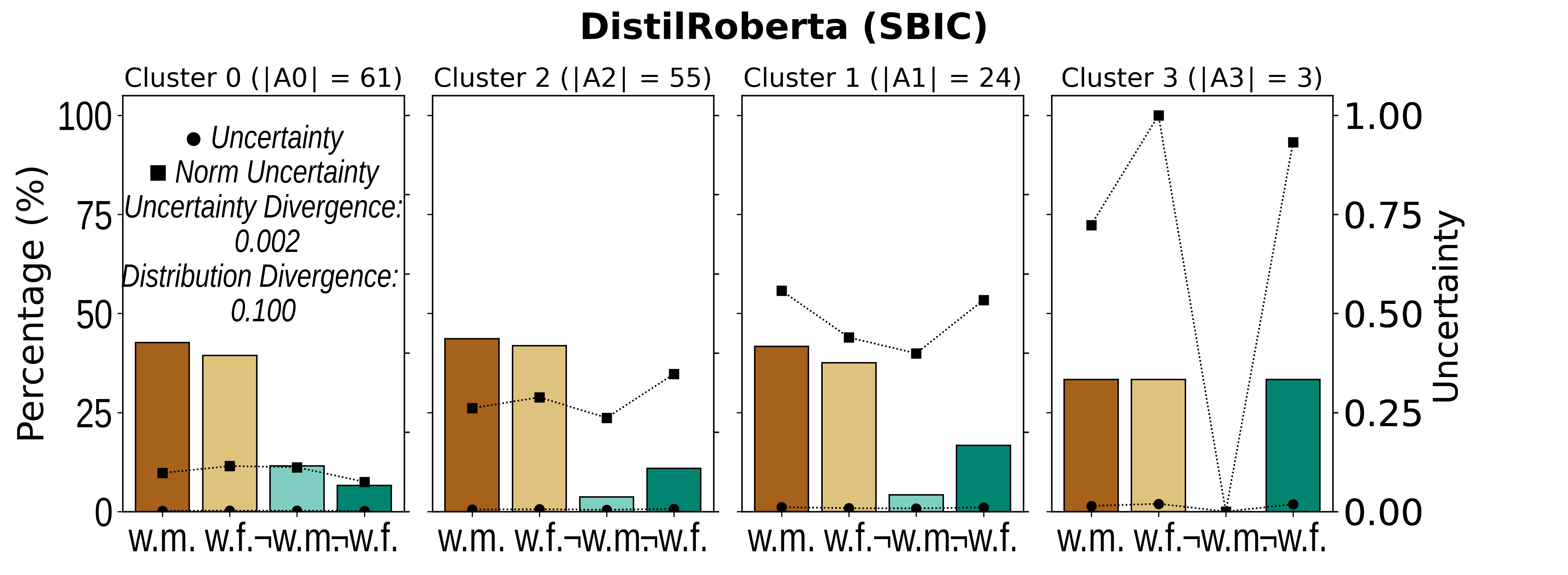}%
        }\\
        \subfloat[]{%
            \includegraphics[width=.48\linewidth]{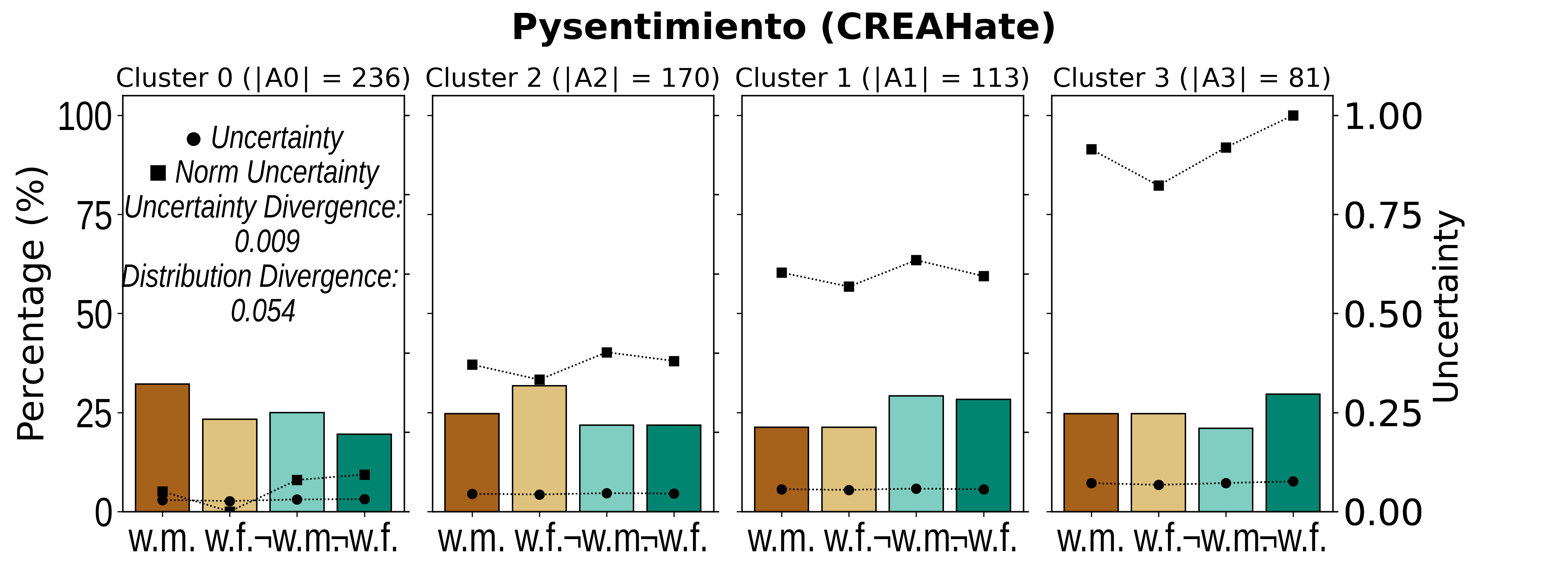}%
        }\hfill
        \subfloat[]{%
            \includegraphics[width=.48\linewidth]{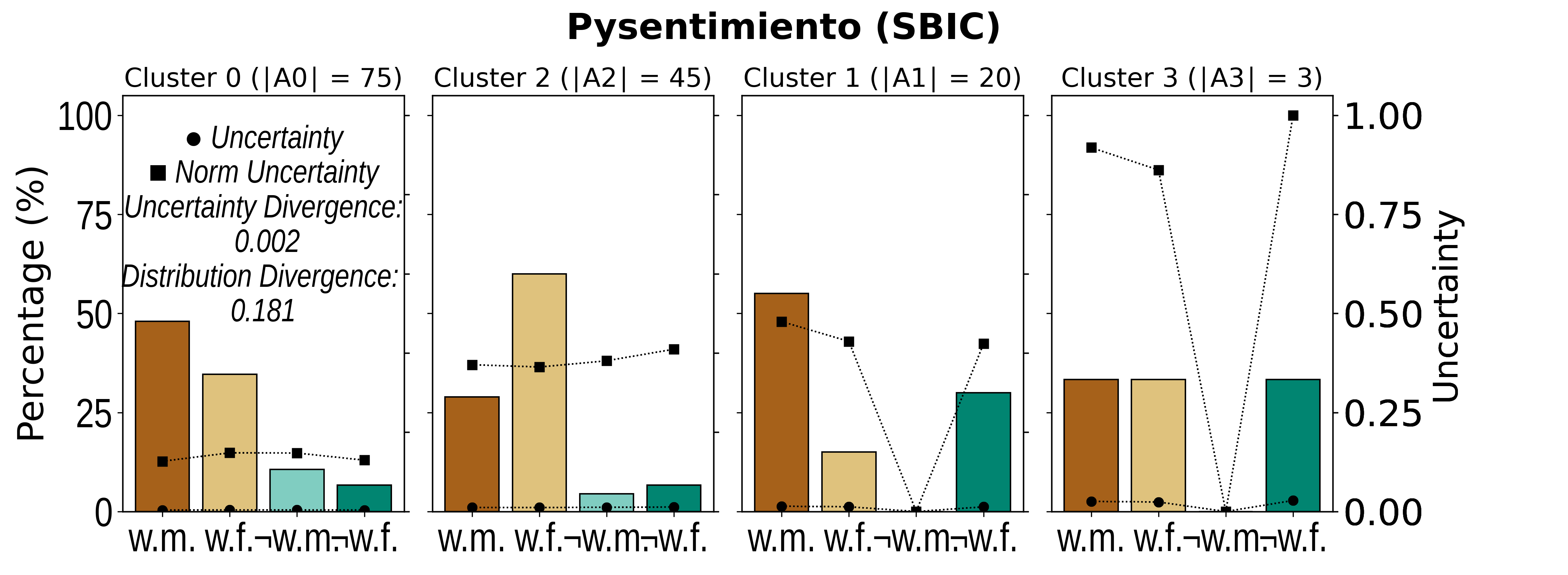}%
        }\\
        \subfloat[]{%
            \includegraphics[width=.48\linewidth]{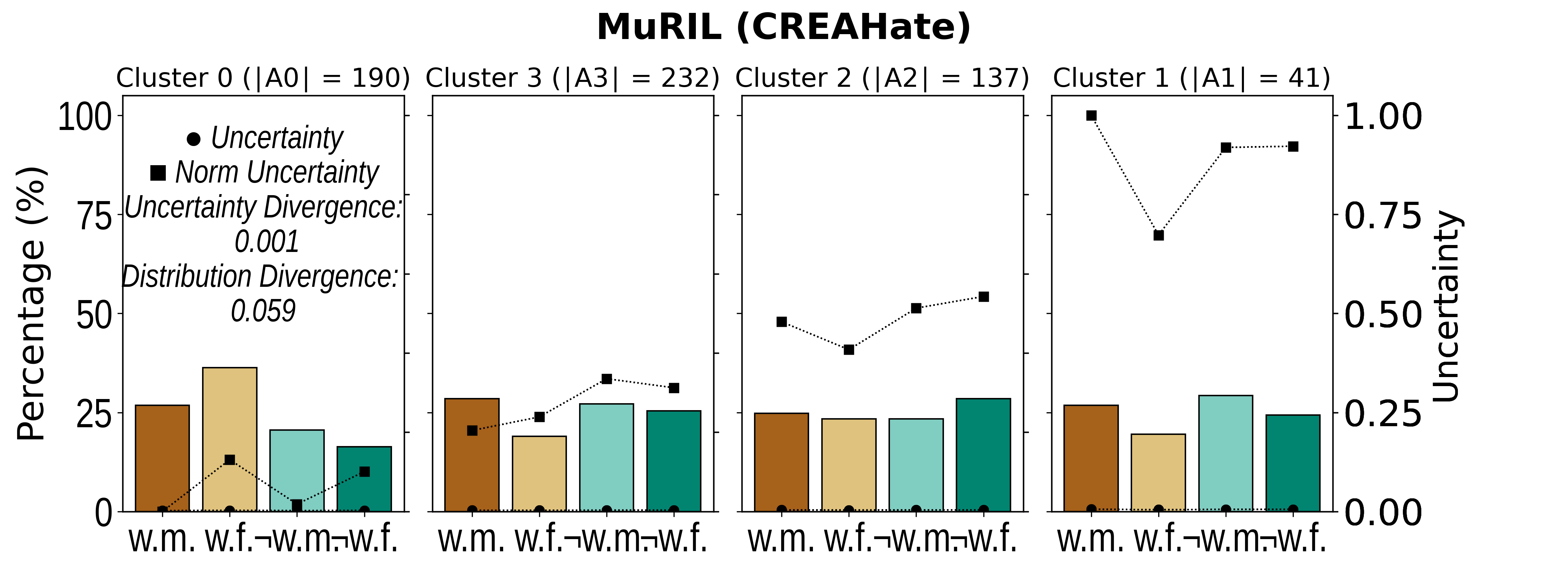}%
        }\hfill
        \subfloat[]{%
            \includegraphics[width=.48\linewidth]{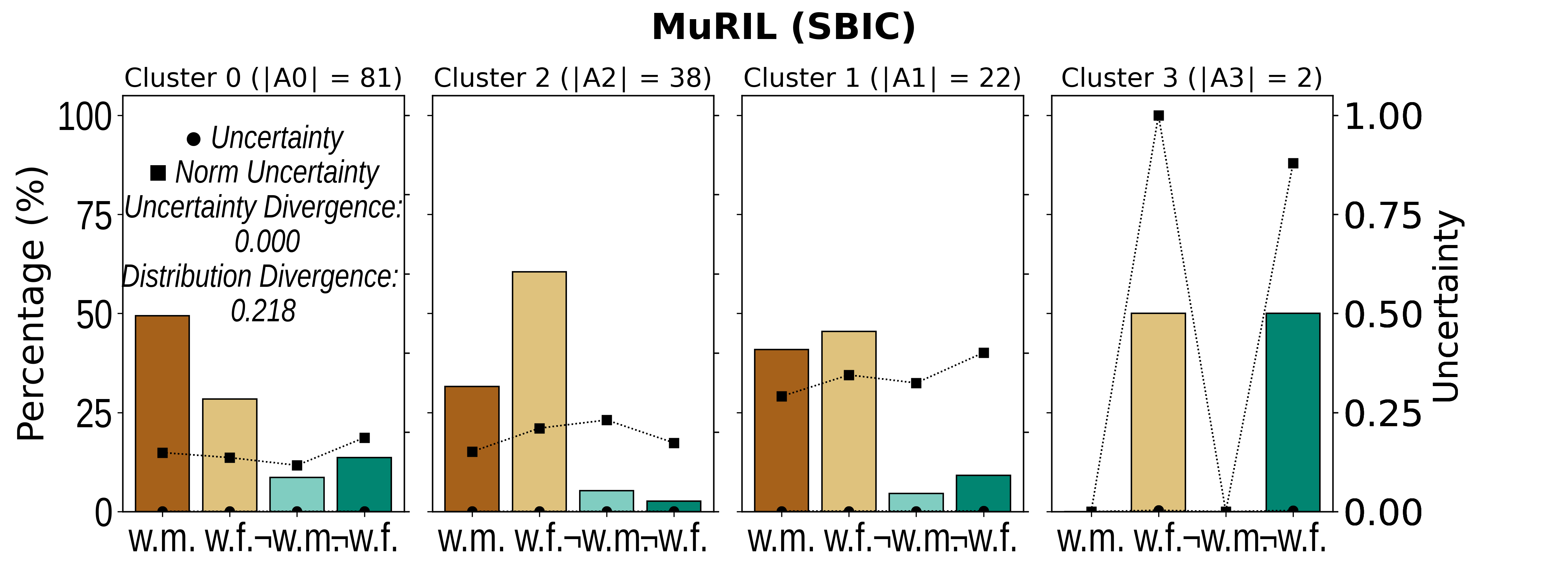}%
        }\\
        \subfloat[]{%
            \includegraphics[width=.48\linewidth]{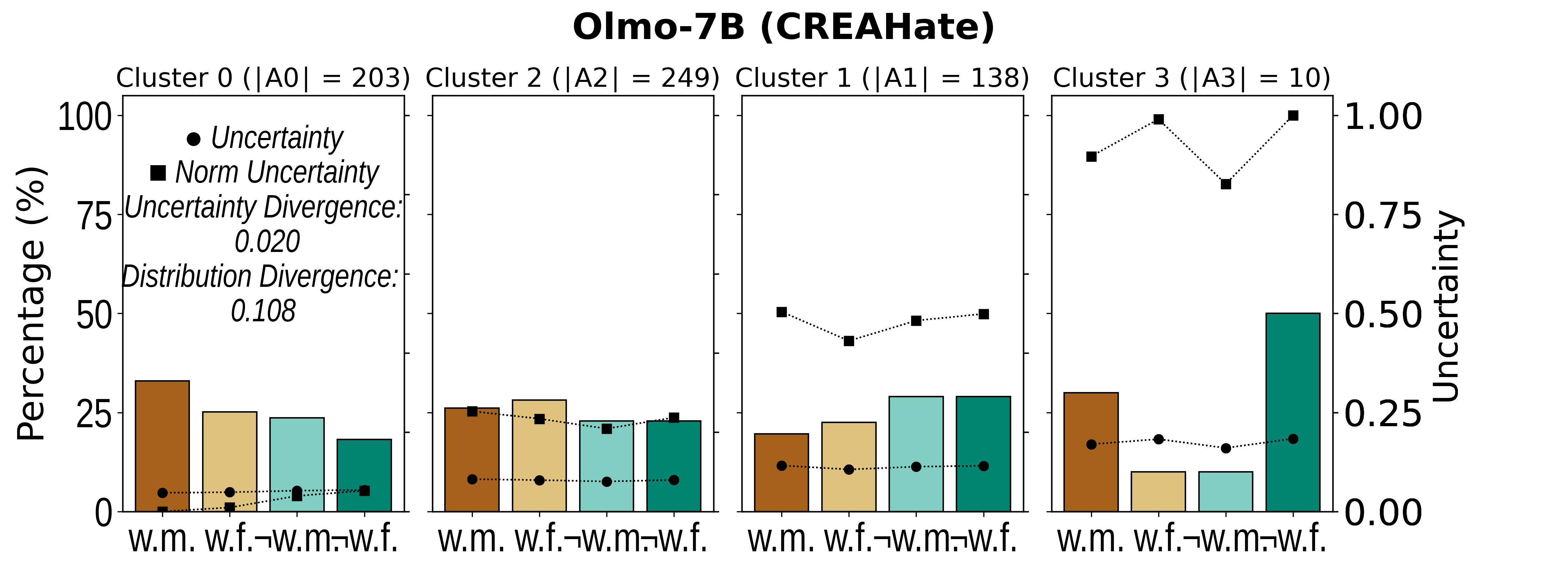}%
        }\hfill
        \subfloat[]{%
            \includegraphics[width=.48\linewidth]{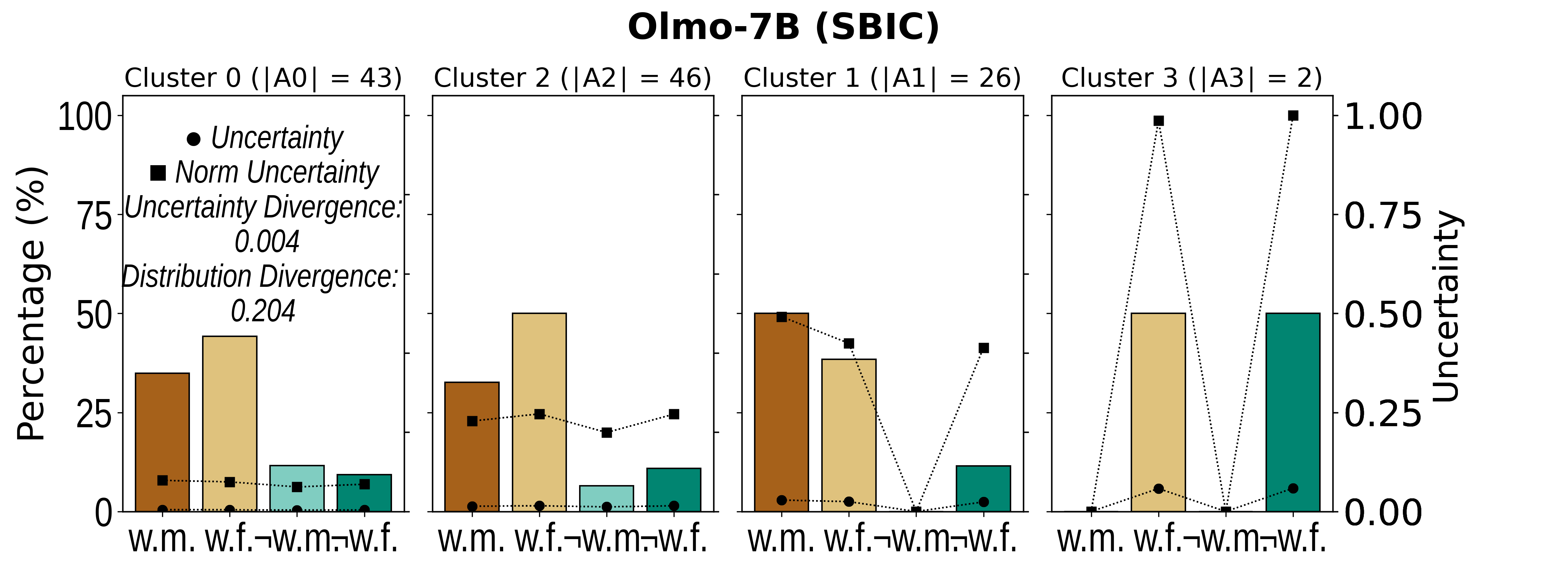}%
        }\\
        \subfloat[]{%
            \includegraphics[width=.48\linewidth]{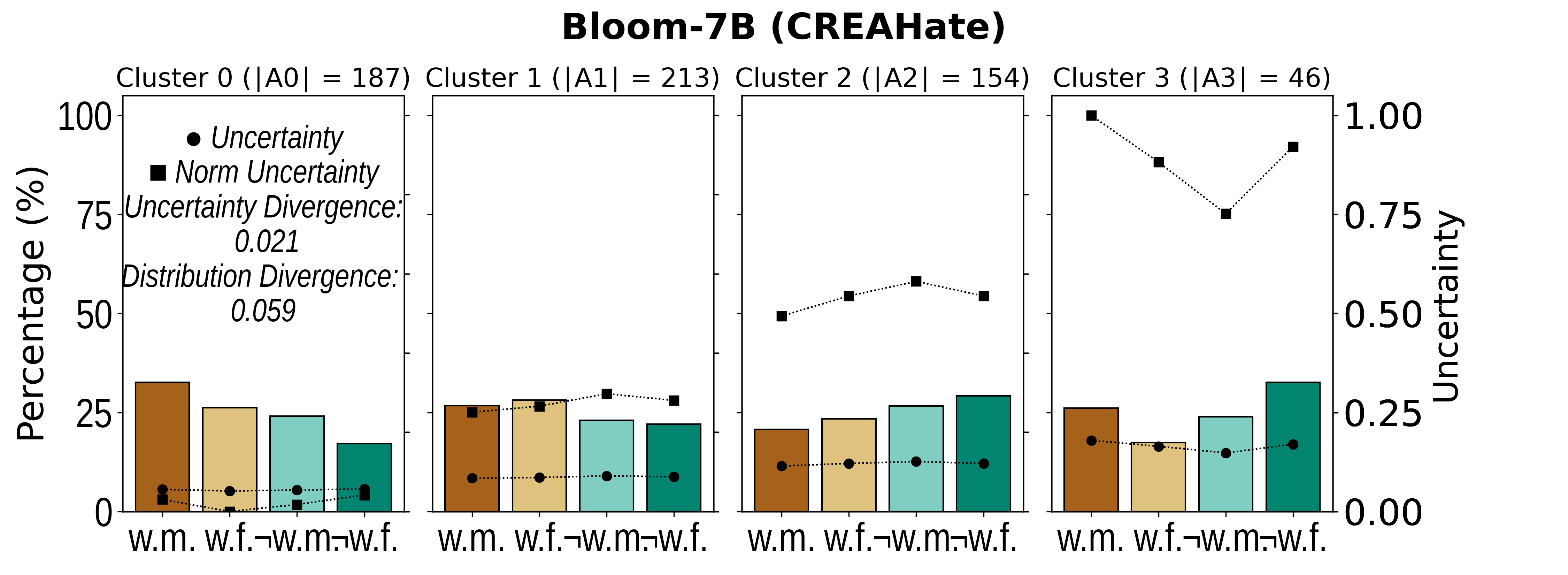}%
        }\hfill
        \subfloat[]{%
            \includegraphics[width=.48\linewidth]{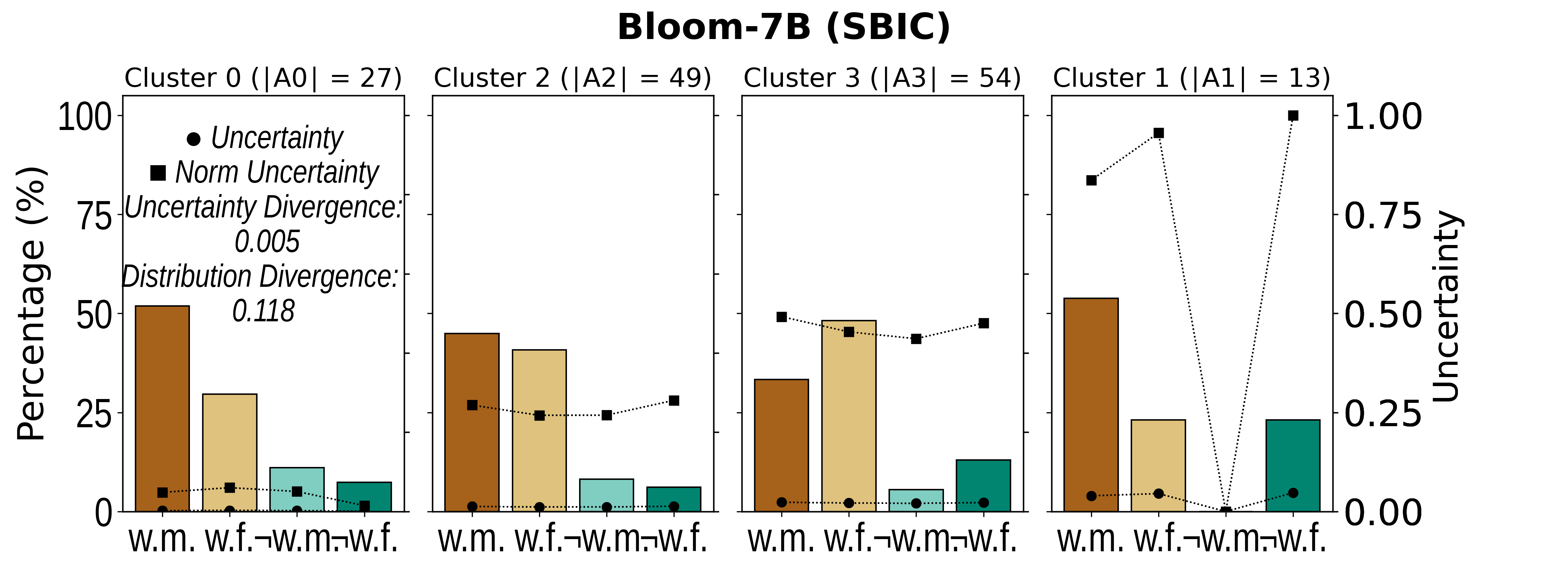}%
        }\\
        \subfloat[]{%
            \includegraphics[width=.48\linewidth]{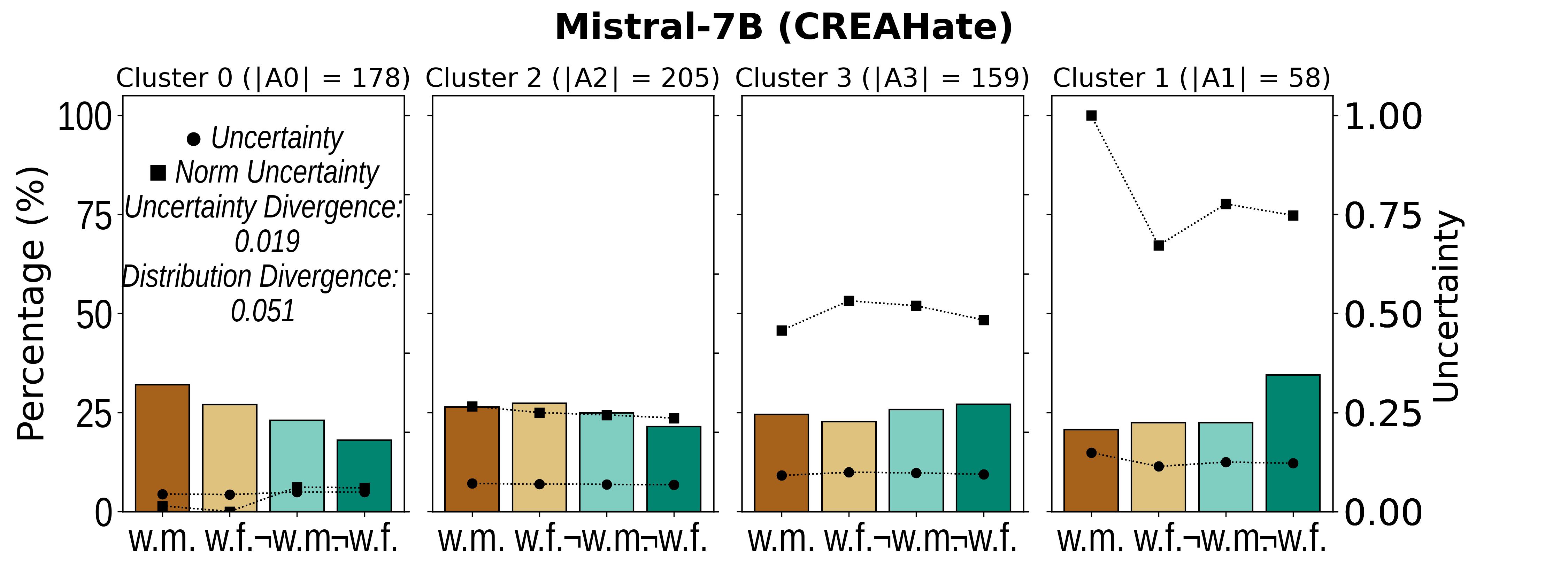}%
        }\hfill
        \subfloat[]{%
            \includegraphics[width=.48\linewidth]{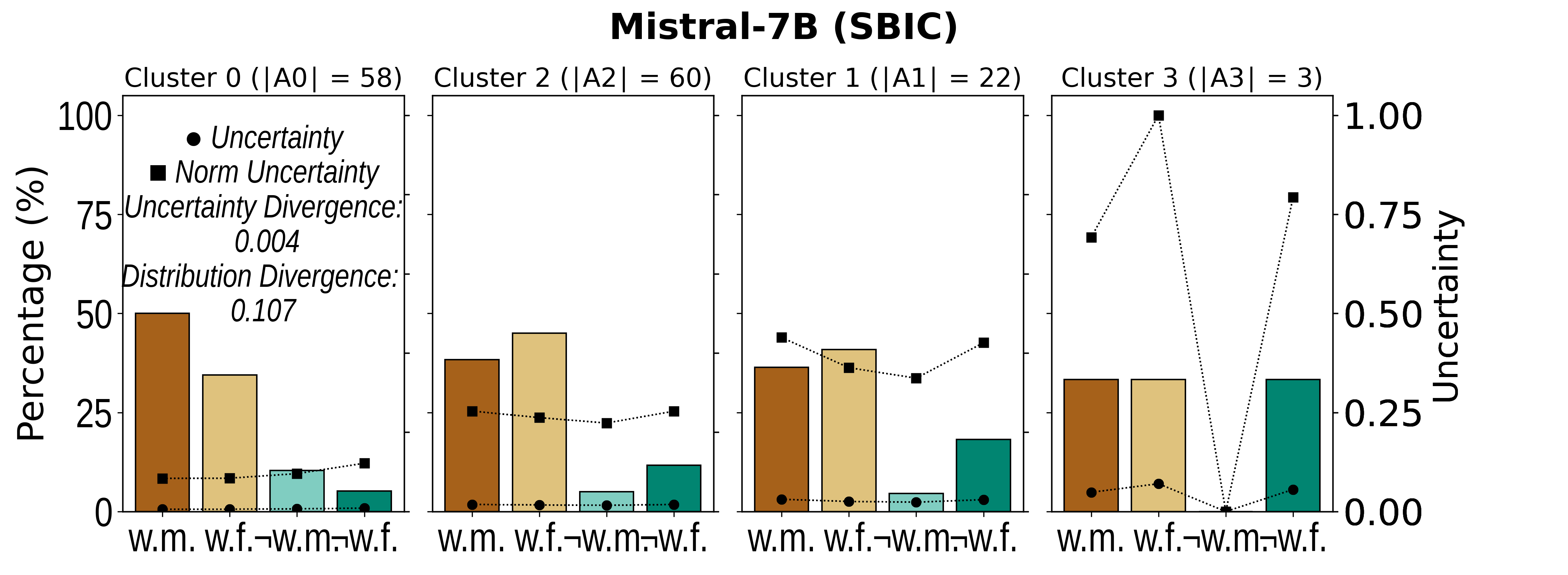}%
        }

        \caption{There are 22 plots, corresponding to the evaluation of 11 models on two distinct datasets, CREAHate and SBIC. Each plot consists of four subplots, where each subplot represents a cluster and illustrates the demographic distribution of the annotators within that cluster. Additionally, the subplots display the level of uncertainty, including its normalized value across the four clusters.}
        \label{fig:appendix}
    \end{figure*}

\section{Cohen's Kappa between annotators pairs}
\label{appendix:appendix_cohen}

The following heatmaps show the pairwise Cohen’s Kappa agreement between annotator classes for the two datasets analyzed in this study (SBIC and CREAHAte). Each matrix displays agreement values between the four demographic groups.

As expected, higher agreement is observed along the diagonal, indicating that annotators within the same class tend to be more consistent in their annotations. Off-diagonal values represent inter-class agreement, which is generally lower, highlighting differences in annotation behavior across demographic groups. These visualizations provide a detailed view of intra- and inter-class consistency and help contextualize the results reported in the main text.

\begin{figure}[htbp]
    \centering
    \includegraphics[width=0.5\textwidth]{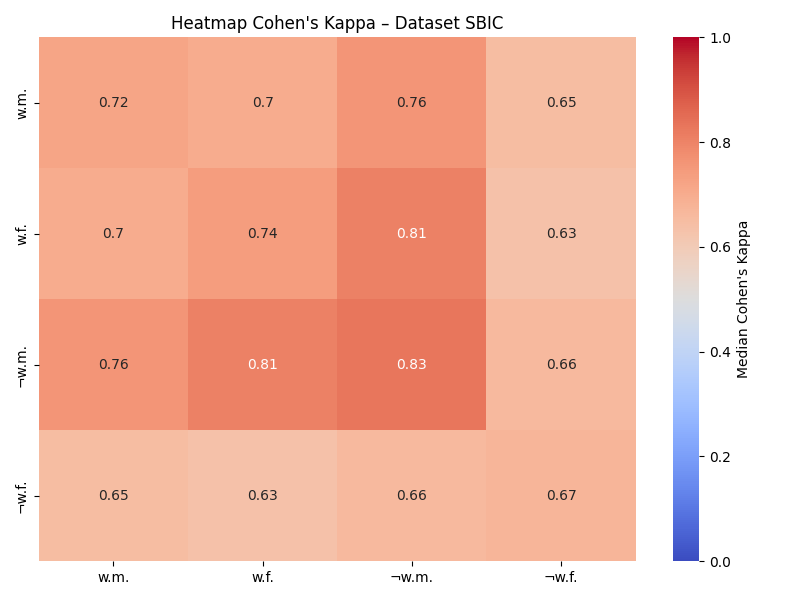} 
    \caption{Pairwise Cohen’s Kappa agreement between annotator classes per SBIC.}
\end{figure}

\begin{figure}[htbp]
    \centering
    \includegraphics[width=0.5\textwidth]{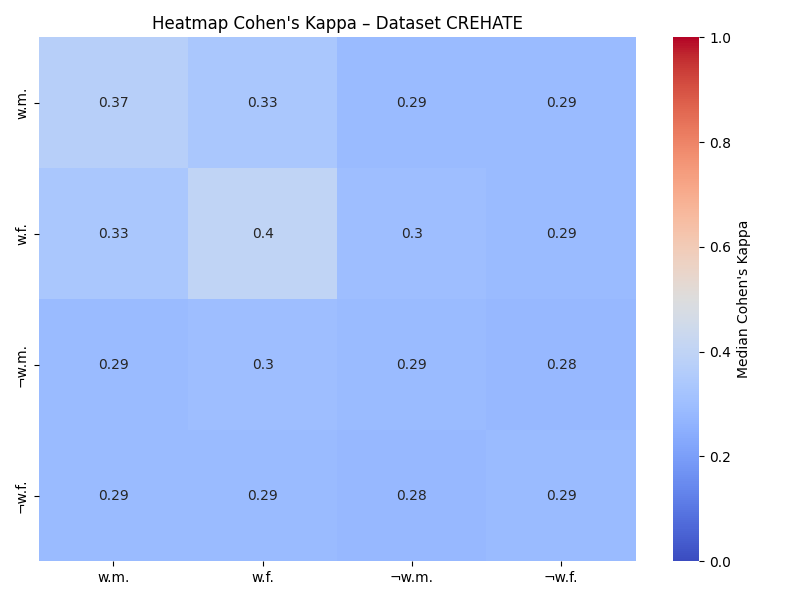} 
    \caption{Pairwise Cohen’s Kappa agreement between annotator classes per CREAHate.}
\end{figure}


\end{document}